\def\year{2019}\relax
\documentclass[letterpaper]{article} 
\usepackage{aaai19}  
\usepackage{times}  
\usepackage{helvet}  
\usepackage{courier}  
\usepackage{url}  
\usepackage{graphicx}  
\usepackage{color}
\usepackage{amssymb}
\usepackage{xspace}
\usepackage{amsmath}
\usepackage{adjustbox}
\usepackage{multirow}

\newcommand{\reluplex}{{\small \textsf{Reluplex}}\xspace}
\newcommand{\cnncert}{{\small \textsf{CNN-Cert}}\xspace}
\newcommand{\cnncertrelu}{{\small \textsf{CNN-Cert-Relu}}\xspace}
\newcommand{\cnncertada}{{\small \textsf{CNN-Cert-Ada}}\xspace}
\newcommand{\fastlin}{{\small \textsf{Fast-Lin}}\xspace}

\newcommand{\crownName}{{\small \textsf{CROWN}}\xspace}

\newcommand{\x}{\mathbf{x}}
\newcommand{\xo}{\mathbf{x_0}}
\newcommand{\W}[1]{\mathbf{W}^{#1}}
\newcommand{\A}[1]{\mathbf{A}^{#1}}

\newcommand{\B}[1]{\mathbf{B}^{#1}}

\newcommand{\upbnd}[1]{\mathbf{u}^{#1}}
\newcommand{\lwbnd}[1]{\mathbf{l}^{#1}}

\newcommand{\bias}[1]{\mathbf{b}^{#1}}

\newcommand{\R}{\mathbb{R}}
\newcommand{\Ball}{\mathbb{B}_{p}(\xo,\epsilon)}
\newcommand{\Ph}[1]{\Phi^{#1}}

\begin{document}
%
\title{CNN-Cert: An Efficient Framework for Certifying Robustness of Convolutional Neural Networks}
\author{Akhilan Boopathy\textsuperscript{1}, Tsui-Wei Weng\textsuperscript{1}, Pin-Yu Chen\textsuperscript{2}, Sijia Liu\textsuperscript{2} and Luca Daniel\textsuperscript{1}\\
\textsuperscript{1}{Massachusetts Institute of Technology, Cambridge, MA 02139} \\
\textsuperscript{2} MIT-IBM Watson AI Lab, IBM Research\\
}
\maketitle
\begin{abstract}
Verifying robustness of neural network classifiers has attracted great interests and attention due to the success of deep neural networks and their unexpected vulnerability to adversarial perturbations. Although finding minimum adversarial distortion of neural networks (with ReLU activations) has been shown to be an NP-complete problem, obtaining a non-trivial lower bound of minimum distortion as a provable robustness guarantee is possible. However, most previous works only focused on simple fully-connected layers (multilayer perceptrons) and were limited to ReLU activations. This motivates us to propose a general and efficient framework, CNN-Cert, that is capable of certifying robustness on general convolutional neural networks. Our framework is general -- we can handle various architectures including convolutional layers, max-pooling layers, batch normalization layer, residual blocks, as well as general activation functions; our approach is efficient -- by exploiting the special structure of convolutional layers, we achieve up to 17 and 11 times of speed-up compared to the state-of-the-art certification algorithms (e.g. Fast-Lin, CROWN) and 366 times of speed-up compared to the dual-LP approach while our algorithm obtains similar or even better verification bounds. In addition, CNN-Cert generalizes state-of-the-art algorithms e.g. Fast-Lin and CROWN. We demonstrate by extensive experiments that our method outperforms state-of-the-art lower-bound-based certification algorithms in terms of both bound quality and speed.    
\end{abstract}

\section{Introduction}

\begin{table*}[tbh!]
\centering
\caption{Comparison of methods for providing adversarial robustness certification in NNs.} 
\label{table_comparison}
\scalebox{0.65}
{
\begin{tabular}{l|cccccc}
\hline  Method  & Non-trivial bound      & Multi-layer  & Scalability \& Efficiency  & Beyond ReLU  &  Exploit CNN structure  & Pooling and other struc.\\ \hline
Reluplex~\cite{katz2017reluplex}, Planet~\cite{ehlers2017formal} & \checkmark & \checkmark & $\times$     & $\times$ & $\times$ & $\times$ \\
Global Lipschitz constant~\cite{szegedy2013intriguing} & $\times$ & \checkmark & \checkmark & \checkmark & $\times$ & \checkmark  \\
Local Lipschitz constant~\cite{hein2017formal}   & \checkmark   & $\times$     & \checkmark & differentiable & $\times$ & $\times$ \\
SDP approach~\cite{raghunathan2018certified} & \checkmark & $\times$     & $\times$     & \checkmark & $\times$ & $\times$\\
Dual approach~\cite{kolter2017provable}      & \checkmark   & \checkmark & \checkmark  & $\times$ & $\times$ & $\times$ \\
Dual approach~\cite{dvijotham2018dual}       & \checkmark   & \checkmark & codes not yet released  & \checkmark & $\times$ & \checkmark     \\
Fast-lin~/~Fast-lip \cite{weng2018towards}   & \checkmark    & \checkmark & \checkmark & $\times$  & $\times$ & $\times$  \\
CROWN~\cite{zhang2018crown}   & \checkmark    & \checkmark & \checkmark & \checkmark  & $\times$ & $\times$ \\
CNN-Cert (This work)                        & \checkmark & \checkmark & \checkmark & \checkmark & \checkmark & \checkmark \\ \hline
\end{tabular}
}
\end{table*}

Recently, studies on adversarial robustness of state-of-the-art machine learning models, particularly neural networks (NNs), have received great attention due to interests in model explainability \cite{goodfellow2014explaining} and rapidly growing concerns on security implications \cite{biggio2017wild}. Take object recognition as a motivating example, imperceptible adversarial perturbations of natural images can be easily crafted to manipulate the model predictions, known as prediction-evasive adversarial attacks. One widely-used threat model to quantify the attack strengths is the norm-ball bounded attacks, where the distortion between an original example and the corresponding adversarial example is measured by the $\ell_p$ norm of their difference in real-valued vector representations (e.g., pixel values for images or embeddings for texts). Popular norm choices are $\ell_1$ \cite{chen2017ead}, $\ell_2$ \cite{carlini2017towards}, and $\ell_\infty$ \cite{kurakin2016adversarial_ICLR}.

The methodology of evaluating model robustness against adversarial attacks can be divided into two categories: \textit{game-based} or \textit{verification-based}.
Game-based approaches measure the success in mitigating adversarial attacks via mounting empirical validation against a (self-chosen) set of attacks. However, many defense methods have shown to be broken or bypassed by attacks that are adaptive to these defenses under the same threat model \cite{carlini2017adversarial,athalye2018obfuscated}, and therefore their robustness claims may not extend to untested attacks.
On the other hand, verification-based approaches provide certified defense against any possible attacks under a threat model. In the case of an $\ell_p$ norm-ball bounded threat model, a verified robustness certificate $\epsilon$ means the (top-1) model prediction on the input data cannot be altered if the attack strength (distortion measured by $\ell_p$ norm) is smaller than $\epsilon$. Different from game-based approaches, verification methods are attack-agnostic and hence can formally certify robustness guarantees, which is crucial to security-sensitive and safety-critical applications.

Although verification-based approaches can provide robustness certification, finding the minimum distortion (i.e., the maximum certifiable robustness) of NNs with ReLU activations has been shown to be an NP-complete problem \cite{katz2017reluplex}. While minimum distortion can be attained in small and shallow networks  \cite{katz2017reluplex,lomuscio2017approach,cheng2017maximum,fischetti2017deep}, these approaches are not even scalable to moderate-sized NNs. Recent works aim to circumvent the scalability issue by efficiently solving a non-trivial lower bound on the minimum distortion \cite{kolter2017provable,weng2018towards,dvijotham2018dual}. However, existing methods may lack generality in supporting different network architectures and activation functions. In addition, current methods often deal with convolutional layers by simply converting back to fully-connected layers, which may lose efficiency if not fully optimized with respect to the NNs, as demonstrated in our experiments. To bridge this gap, we propose \cnncert, a general and efficient verification framework for certifying robustness of a broad range of convolutional neural networks (CNNs). The generality of \cnncert enables robustness certification of various architectures, including convolutional layers, max-pooling layers batch normalization layers and residual blocks, and general activation functions. The efficiency of \cnncert is optimized by exploiting the convolution operation. A full comparison of verification-based methods is given in Table \ref{table_comparison}.

We highlight the contributions of this paper as follows.
\begin{itemize}
    \item \cnncert is \textit{general}  -- it can certify robustness on general CNNs with various building blocks, including convolutional/pooling/batch-norm layers and residual blocks, as well as general activation functions such as ReLU, tanh, sigmoid and arctan. Other variants can easily be incorporated. Moreover, certification algorithms Fast-Lin~\cite{weng2018towards} and CROWN~\cite{zhang2018crown} are special cases of \cnncert.  
    \item \cnncert is \textit{computationally efficient} -- the cost is similar to forward-propagation as opposed to NP-completeness in formal verification methods, e.g. \reluplex \cite{katz2017reluplex}. Extensive experiments show that \cnncert achieves up to 17 times of speed-up compared to state-of-the-art certification algorithms \fastlin and up to 366 times of speed-up compared to dual-LP approaches while \cnncert obtains similar or even better verification bounds.
\end{itemize}

\vspace{-2.5mm}

\section{Background and Related Work}

\paragraph{Adversarial Attacks and Defenses.}
In the white-box setting where the target model is entirely transparent to an adversary, recent works have demonstrated adversarial attacks on machine learning applications empowered by neural networks, including object recognition \cite{szegedy2013intriguing}, image captioning \cite{chen2017show}, machine translation \cite{cheng2018seq2sick}, and graph learning \cite{zugner2018adversarial}. Even worse, adversarial attacks are still plausible in the black-box setting, where the adversary is only allowed to access the model output but not the model internals \cite{CPY17zoo,ilyas2018black,tu2018autozoom,cheng2018query}.
For improving the robustness of NNs, adversarial training with adversarial attacks is by far one of the most effective strategies that showed strong empirical defense performance~\cite{madry2017towards,sinha2017certifiable}. In addition, verification-based methods have validated that NNs with adversarial training can indeed improve robustness~\cite{kolter2017provable,weng2018evaluating}. 

\vspace{-2.5mm}

\paragraph{Robustness Verification for Neural Networks.}
Under the norm-ball bounded threat model, for NNs with ReLU activation functions, although the minimum adversarial distortion gives the best possible certified robustness, solving it is indeed computationally intractable due to its NP-completeness complexity \cite{katz2017reluplex}. Alternatively, solving a non-trivial lower bound of the minimum distortion as a provable robustness certificate is a more promising option but at the cost of obtaining a more conservative robustness certificate. Some analytical lower bounds depending solely on model weights can be derived \cite{szegedy2013intriguing,peck2017lower,hein2017formal,raghunathan2018certified} but 
they are in general too loose to be useful or limited to 1 or 2 hidden layers. The robustness of NNs can be efficiently certified on ReLU activation~\cite{kolter2017provable,weng2018towards} and general activation~\cite{zhang2018crown} but mostly on models with fully-connected layers. \cite{dvijotham2018dual} can also be applied to different activation functions but their bound quality might decrease a lot as a trade-off between computational efficiency due to its `any-time` property. This paper falls within this line of research with an aim of providing both a general and efficient certification framework for CNNs (see Table \ref{table_comparison} for detailed comparisons).

\vspace{-3mm}

\paragraph{Threat model, minimum adversarial distortion $\rho_{\text{min}}$ and certified lower bound $\rho_{\text{cert}}$.} 
Throughout this paper, we consider the $\ell_p$ norm-ball bounded threat model with full access to all the model parameters.
Given an input image $\xo$ and a neural network classifier $f(\x)$, let $c = \text{argmax}_{i} f_i(\xo)$ be the class where $f$ predicts for $\xo$. The minimum distortion $\rho_{\text{min}}$ is the smallest perturbation that results in $\text{argmax}_{i} f_i(\xo + \delta) \neq c$, and $\rho_{\text{min}} = \|\delta\|_p$. A certified lower bound $\rho_{\text{cert}}$ satisfies the following: (i) $\rho_{\text{cert}} < \rho_{\text{min}}$ and (ii) for all $\delta \in \R^d$ and $\| \delta \|_p \leq \rho_{\text{cert}}$, $\text{argmax}_{i} f_i(\xo + \delta) = c$. In other words, a certified bound \textit{guarantees} a region (an $\ell_p$ ball with radius $\rho_{\text{cert}}$) such that the classifier decision can never be altered for all possible perturbations in that region. Note that $\rho_{\text{cert}}$ is also known as \textit{un-targeted} robustness, and the \textit{targeted} robustness $\rho_{\text{cert,t}}$ is defined as satisfying (i) but with (ii) slightly modified as $\forall \delta \in \R^d$ and $\| \delta \|_p \leq \rho_{\text{cert}}$, $f_c(\xo + \delta) > f_t(\xo + \delta)$ given some targeted class $t \neq c$.




\newtheorem{theorem}{Theorem}[section]
\newtheorem{lemma}[theorem]{Lemma}
\newtheorem{definition}[theorem]{Definition}
\newtheorem{notation}[theorem]{Notation}
\newtheorem{proposition}[theorem]{Proposition}
\newtheorem{corollary}[theorem]{Corollary}
\newtheorem{conjecture}[theorem]{Conjecture}
\newtheorem{assumption}[theorem]{Assumption}
\newtheorem{observation}[theorem]{Observation}
\newtheorem{fact}[theorem]{Fact}
\newtheorem{remark}[theorem]{Remark}
\newtheorem{claim}[theorem]{Claim}
\newtheorem{example}[theorem]{Example}
\newtheorem{problem}[theorem]{Problem}
\newtheorem{open}[theorem]{Open Problem}
\newtheorem{property}[theorem]{Property}
\newtheorem{hypothesis}[theorem]{Hypothesis}
\definecolor{Sijia_color}{rgb}{0.858, 0.188, 0.478}

\section{CNN-Cert: A General and Efficient Framework for Robustness Certification} \label{sec:theory}

\begin{table*}[t]
\centering
\caption{Expression of $\A{r}_{U}$ and $\B{r}_{U}$. $\A{r}_{L}$ and $\B{r}_{L}$ have exactly the same form as $\A{r}_{U}$ and $\B{r}_{U}$ but with $U$ and $L$ swapped.}
\label{tab:ABexpression}
\begin{tabular}{l|c|c}
\hline
Blocks           & $\A{r}_{U,(\vec{x},z),(\vec{i},k)}$  & $\B{r}_{U}$   \\ \hline
(i) Act-Conv Block  & {\small $\W{r+}_{(\vec{x},z),(\vec{i},k)}\alpha_{U,(\vec{i}+\vec{x},k)} + \W{r-}_{(\vec{x},z),(\vec{i},k)}\alpha_{L,(\vec{i}+\vec{x},k)}$}  & {\small $\W{r+}*(\alpha_{U} \odot \beta_{U}) + \W{r-}*(\alpha_{L} \odot \beta_{L}) + \bias{r}$}  \\
(ii) Residual Block & {\small $[\A{r}_{U,\text{act}}*\W{r-1} + I]_{(\vec{i},k),(\vec{x},z)}$} &  {\small $\A{r}_{U,\text{act}}*\bias{r-1}+\B{r}_{U,\text{act}}$} \\
(iv) Pooling Block &  $\frac{\upbnd{}_{(\vec{i}+\vec{x},z)}-\gamma}{\upbnd{}_{(\vec{i}+\vec{x},z)}-\lwbnd{}_{(\vec{i}+\vec{x},z)}}$ & at location $(\vec{x},z)$: $\sum_{\vec{i} \in S_n} \frac{(\gamma-\upbnd{}_{(\vec{i}+\vec{x},z)})\lwbnd{}_{(\vec{i}+\vec{x},z)}}{\upbnd{}_{(\vec{i}+\vec{x},z)}-\lwbnd{}_{(\vec{i}+\vec{x},z)}} + \gamma$ \\ 
& {\small $\gamma = \min \{ \max \{ \gamma_0, \max \lwbnd{}_{S} \}, \min \upbnd{}_{S} \}$} & {\small $\gamma_0 = \frac{\sum_{S} \frac{\upbnd{}_{S}}{\upbnd{}_{S}-\lwbnd{}_{S}}-1}{\sum_{S} \frac{1}{\upbnd{}_{S}-\lwbnd{}_{S}}}$}  \\
\hline \hline
\multicolumn{3}{l}{{\small Note 1: $(\vec{i},k) = (i,j,k)$ denotes filter coordinate indices and $(\vec{x},z) = (x,y,z)$ denotes output tensor indices.}} \\
\multicolumn{3}{l}{{\small Note 2: $\A{r}_{U},\B{r}_{U},\W{},\alpha,\beta,\upbnd{},\lwbnd{}$ are all tensors. $\W{r+},\W{r-}$ contains only the positive, negative entries of $\W{r}$ with other entries equal 0.}} \\
\multicolumn{3}{l}{{\small Note 3: $\A{r}_{L},\B{r}_{L}$ for pooling block are slightly different. Please see Appendix (c) for details.}}\\
\hline
\end{tabular}
\end{table*}

\paragraph{Overview of our results.}
In this section, we present a general and efficient framework \cnncert for computing certified lower bounds of minimum adversarial distortion with general activation functions in CNNs. We derive the range of network output in closed-form by applying a pair of linear upper/lower bound on the neurons (e.g. the activation functions, the pooling functions) when the input of the network is perturbed with noises bounded in $\ell_p$ norm ($p \geq 1$). Our framework can incorporate general activation functions and various architectures -- particularly, we provide results on convolutional layers with activations (a.k.a Act-conv block), max-pooling layers (a.k.a. Pooling block), residual blocks (a.k.a. Residual block) and batch normalization layers (a.k.a. BN block). In addition, we show that the state-of-the-art \fastlin algorithm \cite{weng2018towards} and \crownName~\cite{zhang2018crown} are special cases under the \cnncert framework. 

\subsection{General framework}\label{sec:framework}
\begin{figure*}[h]
\centering
\includegraphics[width=0.9\textwidth]{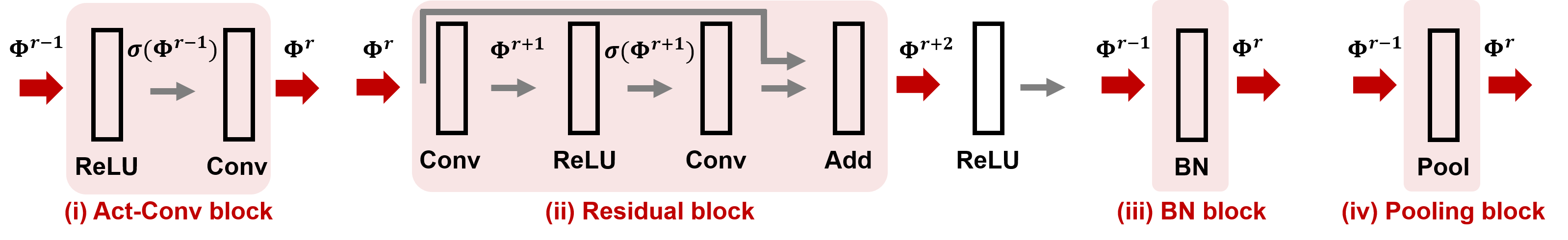}
\caption{Cartoon graph of commonly-used building blocks (i)-(iv) considered in our \cnncert framework. The key step in deriving explicit network output bound is to consider the input/output relations of each building block, marked as red arrows.  The activation layer can be general activations but here is denoted as ReLU.}
\label{fig:diagram}
\end{figure*}

When an input data point is perturbed within an $\ell_p$ ball with radius $\epsilon$, we are interested in the change of network output because this information can be used to find a certified lower bound of minimum adversarial distortion (as discussed in the section \textbf{Computing certified lower bound $\rho_{\text{cert}}$}). Toward this goal, the first step is to derive explicit output bounds for the neural network classifiers with various popular building blocks, as shown in Figure~\ref{fig:diagram}, Table~\ref{tab:ABexpression} and Table \ref{tab:ABexpression1} (with general strides and padding). The fundamental idea of our method is to apply \textit{linear} bounding techniques separately on the \textit{non-linear} operations in the neural networks, e.g. the non-linear activation functions, residual blocks and pooling operations. Our proposed techniques are general and allow efficient computations of certified lower bounds. We begin the formal introduction to \cnncert by giving notations and intuitions of deriving explicit bounds for each building block followed by the descriptions of utilizing such explicit bounds to compute certified lower bounds $\rho_{\text{cert}}$ in our proposed framework. 
\vspace{-3.5mm}

\paragraph{Notations.} Let $f(\x)$ be a neural network classifier function and $\xo$ be an input data point. We use $\sigma(\cdot)$ to denote the coordinate-wise activation function in the neural networks.  Some popular choices of $\sigma$ include ReLU: $\sigma(y) = \max(y,0)$, hyperbolic tangent: $\sigma(y) = \tanh(y)$, sigmoid: $\sigma(y) = 1/(1+e^{-y})$ and arctan: $\sigma(y) = \tan^{-1}(y)$. The symbol $*$ denotes the convolution operation and $\Ph{r}(\x)$ denotes the output of $r$-th layer building block, which is a function of an input $\x$. We use superscripts to denote index of layers and subscripts to denote upper bound ($U$), lower bound ($L$) and its corresponding building blocks (e.g. act is short for activation, conv is short for convolution, res is short for residual block, bn is short for batch normalization and pool is short for pooling). Sometimes subscripts are also used to indicate the element index in a vector/tensor, which is self-content. We will often write $\Ph{r}(\x)$ as $\Ph{r}$ for simplicity and we will sometimes use $\Ph{m}(\x)$ to denote the output of the classifier, i.e. $\Ph{m} = f(\x)$. Note that the weights $\W{}$, bias $\bias{}$, input $\x$ and the output $\Ph{m}$ of each layer are tensors since we consider a general CNN in this paper. 
\vspace{-3.5mm}

\paragraph{(i) Tackling the non-linear activation functions and convolutional layer.} 
For the convolutional layer with an activation function $\sigma(\cdot)$, let $\Ph{r-1}$ be the input of activation layer and $\Ph{r}$ be the output of convolutional layer. The input/output relation is as follows:
\begin{equation}
\label{eq:conv_relation}
    \Ph{r} = \W{r}*\sigma(\Ph{r-1}) + \bias{r}.
\end{equation}
Given the range of $\Ph{r-1}$, we can bound the range of $\Ph{r}$ by applying two linear bounds on each activation function $\sigma(y)$:
\begin{equation}
\label{eq:conv_idea}
    \alpha_L(y+\beta_L) \leq \sigma(y) \leq \alpha_U(y+\beta_U).
\end{equation}
When the input $y$ is in the range of $[l,u]$, the parameters $\alpha_L,\alpha_U,\beta_L,\beta_U$ can be chosen appropriately based on $y$'s lower bound $l$ and upper bound $u$. If we use \eqref{eq:conv_idea} and consider the signs of the weights associated with the activation functions, it is possible to show that the output $\Ph{r}$ in \eqref{eq:conv_relation} can be bounded as follows:
\begin{align}
    \Ph{r} & \leq \A{r}_{U,\text{act}} * \Ph{r-1} + \B{r}_{U,\text{act}} \label{eq:act_ub}, \\
    \Ph{r} & \geq \A{r}_{L,\text{act}} * \Ph{r-1} + \B{r}_{L,\text{act}}, \label{eq:act_lb}
\end{align}
where $\A{r}_{U,\text{act}}, \A{r}_{L,\text{act}}, \B{r}_{U,\text{act}}, \B{r}_{L,\text{act}}$ are constant tensors related to weights $\W{r}$ and bias $\bias{r}$ as well as the corresponding parameters $\alpha_L,\alpha_U,\beta_L,\beta_U$ in the linear bounds of each neuron. See Table~\ref{tab:ABexpression} for full results. Note the bounds in \eqref{eq:act_ub} and \eqref{eq:act_lb} are element-wise inequalities and we leave the derivations in the Appendix (a). On the other hand, if $\Ph{r-1}$ is also the output of convolutional layer, i.e. 
\begin{equation*}
    \Ph{r-1} = \W{r-1}*\sigma(\Ph{r-2}) + \bias{r-1},
\end{equation*}
thus the bounds in \eqref{eq:act_ub} and \eqref{eq:act_lb} can be rewritten as follows:
\begin{align}
    \Ph{r} 
    & \leq \A{r}_{U,\text{act}} * \Ph{r-1} + \B{r}_{U,\text{act}} \nonumber \\ 
    & =  \A{r}_{U,\text{act}} * (\W{r-1}*\sigma(\Ph{r-2}) + \bias{r-1}) + \B{r}_{U,\text{act}} \nonumber \\
    & =  \A{r-1}_{U,\text{conv}} * \sigma(\Ph{r-2}) + \B{r-1}_{U,\text{conv}} + \B{r}_{U,\text{act}} \label{eq:conv_ub}
\end{align}
and similarly
\begin{align}
    \Ph{r} 
    & \geq \A{r}_{L,\text{act}} * \Ph{r-1} + \B{r}_{L,\text{act}} \nonumber \\
    & =  \A{r-1}_{L,\text{conv}} * \sigma(\Ph{r-2}) + \B{r-1}_{L,\text{conv}} + \B{r}_{L,\text{act}} \label{eq:conv_lb}
\end{align}
by letting $\A{r-1}_{U,\text{conv}} = \A{r}_{U,\text{act}} * \W{r-1}$, $\B{r-1}_{U,\text{conv}} = \A{r}_{U,\text{act}} * \bias{r-1}$, and $\A{r-1}_{L,\text{conv}} = \A{r}_{L,\text{act}} * \W{r-1}$, $\B{r-1}_{L,\text{conv}} = \A{r}_{L,\text{act}} * \bias{r-1}$. Observe that the form of the upper bound in \eqref{eq:conv_ub} and lower bound in \eqref{eq:conv_lb} becomes the same convolution form again as \eqref{eq:conv_relation}. Therefore, for a neural network consists of \textit{convolutional} layers and \textit{activation} layers, the above technique can be used iteratively to obtain the final upper and lower bounds of the output $\Ph{r}$ in terms of the input of neural network $\Ph{0}(\x) = \x$ in the following convolutional form:  
\begin{equation*}
    \A{0}_{L, \text{conv}} * \x + \B{0}_L \leq \Ph{r}(\x) \leq \A{0}_{U, \text{conv}} * \x + \B{0}_U.
\end{equation*}
In fact, the above framework is very general and is not limited to the \textit{convolution-activation} building blocks. The framework can also incorporate popular residual blocks, pooling layers and batch normalization layers, etc. The key idea is to derive linear upper bounds and lower bounds for each building block in the form of \eqref{eq:act_ub} and \eqref{eq:act_lb}, and then plug in the corresponding bounds and \textit{back-propagate} to the previous layer. 
\vspace{-3.5mm}

\paragraph{(ii) Tackling the residual blocks operations.}
For the residual block, let $\Ph{r+2}$ denote the output of residual block (before activation) and $\Ph{r+1}$ be the output of first convolutional layer and $\Ph{r}$ be the input of residual block. The input/output relation is as follows:
\begin{align}
    & \Ph{r+1} = \W{r+1}*\Ph{r} + \bias{r+1}, \nonumber \\
    & \Ph{r+2} = \W{r+2}*\sigma(\Ph{r+1}) + \bias{r+2}+ \Ph{r}. \nonumber
\end{align} 
Similar to the linear bounding techniques for up-wrapping the non-linear activation functions, the output of residual block can be bounded as: 
\begin{align}
    \Ph{r+2} &\leq \A{r+2}_{U,\text{res}} * \Ph{r} + \B{r+2}_{U,\text{res}}, \nonumber \\
    \Ph{r+2} &\geq \A{r+2}_{L,\text{res}} * \Ph{r} + \B{r+2}_{L,\text{res}}, \nonumber
\end{align}
where $\A{r+2}_{U,\text{res}}, \A{r+2}_{L,\text{res}}, \B{r+2}_{U,\text{res}}, \B{r+2}_{L,\text{res}}$ are constant tensors related to weights $\W{r+2}$, $\W{r+1}$, bias $\bias{r+2}$, $\bias{r+1}$, and the corresponding parameters $\alpha_L,\alpha_U,\beta_L,\beta_U$ in the linear bounds of each neuron; see Table~\ref{tab:ABexpression} for details. Note that in Table~\ref{tab:ABexpression}, all indices are shifted from $r+2$ to $r$. The full derivations are provided in the Appendix (b).
\vspace{-3.5mm}

\paragraph{(iii) Tackling the batch normalization.} 
The batch normalization layer performs operations of scaling and shifting during inference time. Let $\Ph{r}$ be the output and $\Ph{r-1}$ be the input, the input/output relation is the following:  
\begin{equation*}
    \Ph{r} = \gamma_{\text{bn}} \frac{\Ph{r-1}-\mu_{\text{bn}}}{\sqrt{\sigma_{\text{bn}}^2 + \epsilon_{\text{bn}}}} + \beta_{\text{bn}},
\end{equation*}
where $\gamma_{\text{bn}}$, $\beta_{\text{bn}}$ are the learned training parameters and $\mu_{\text{bn}}$, $\sigma_{\text{bn}}^2$ are the running average of the batch mean and variance during training. Thus, it is simply scaling and shifting on both upper bounds and lower bounds:
\begin{equation*}
    \A{r}_{L, \text{bn}}*\Ph{r-1}+\B{r}_{L, \text{bn}} \leq \Ph{r} \leq  \A{r}_{U, \text{bn}}*\Ph{r-1}+\B{r}_{U, \text{bn}}
\end{equation*}
where $\A{r}_{U, \text{bn}} = \A{r}_{L, \text{bn}} = \frac{\gamma_{\text{bn}}}{\sqrt{\sigma_{\text{bn}}^2 + \epsilon_{\text{bn}}}}$ and $\B{r}_{U, \text{bn}} = \B{r}_{L, \text{bn}} =  -\gamma_{\text{bn}} \frac{\mu_{\text{bn}}}{\sqrt{\sigma_{\text{bn}}^2 + \epsilon_{\text{bn}}}} + \beta_{\text{bn}}$.

\vspace{-3.5mm}

\paragraph{(iv) Tackling the pooling operations.} Let $\Ph{r}$ and  $\Ph{r-1}$ be the output and input of the pooling layer. For max-pooling operations, the input/output relation is the following:
\begin{equation*}
    \Ph{r}_n = \max_{S_n} \; \Ph{r-1}_{S_n},
\end{equation*}
where $S_n$ denotes the pooled input index set associated with the $n$-th output. When the input $\Ph{r-1}$ is bounded in the range $[\lwbnd{r}, \upbnd{r}]$, it is possible to bound the output $\Ph{r}$ by linear functions as follows:
\begin{align}
    \Ph{r} &\leq  \A{r}_{U,\text{pool}} * \Ph{r-1} + \B{r}_{U,\text{pool}}, \nonumber \\
     \Ph{r} &\geq  \A{r}_{L,\text{pool}} * \Ph{r-1} + \B{r}_{L,\text{pool}}, \nonumber
\end{align}
where $\A{r}_{U,\text{pool}}, \A{r}_{L,\text{pool}}, \B{r}_{U,\text{pool}},\B{r}_{L,\text{pool}}$ are constant tensors related to $\lwbnd{r}$ and $\upbnd{r}$. For average pooling operation, the range of the output $\Ph{r}$ is simply the the average of $\lwbnd{r}$ and $\upbnd{r}$ on the corresponding pooling indices. See Table~\ref{tab:ABexpression} and derivation details in Appendix (c).
\vspace{-3.5mm}

\paragraph{Computing global bounds $\eta_{j,U}$ and $\eta_{j,L}$ of network output $\Ph{m}(\x)$.}
Let $\Ph{m}(\x)$ be the output of a $m$-th layer neural network classifier. We have shown that when the input of each building block is bounded and lies in the range of some $[\lwbnd{},\upbnd{}]$, then the output of the building block can be bounded by two linear functions in the form of input convolution. Since a neural network can be regarded as a cascade of building blocks -- the input of current building block is the output of previous building block -- we can propagate the bounds from the last building block that relates the network output \textit{backward} to the first building block that relates the network input $\x$. A final upper bound and lower bound connect the network output and input are in the following linear relationship:
\begin{equation}
\label{eq:form_output}
 \A{0}_L* \x + \B{0}_L  \leq  \Ph{m}(\x) \leq \A{0}_U* \x + \B{0}_U.
\end{equation}
Recall that the input $\x$ is constrained within an $\ell_p$ ball $\Ball$ centered at input data point $\xo$ and with radius $\epsilon$. Thus, maximizing (minimizing) the right-hand side (left-hand side) of \eqref{eq:form_output} over $\x \in \Ball$ leads to a global upper (lower) bound of $j$-th output $\Ph{m}_j(\x)$:
\begin{align}
    \eta_{j,U}  &= \epsilon \| \text{vec}(\A{0}_U) \|_q + \A{0}_U* \xo + \B{0}_U, \label{eq:global_ub} \\
    \eta_{j,L}  &= -\epsilon \| \text{vec}(\A{0}_L) \|_q + \A{0}_L* \xo + \B{0}_L, \label{eq:global_lb}
\end{align}
where $\|\cdot\|_q$ is $\ell_q$ norm and $1/p+1/q = 1$ with $\,p, q\geq 1$. 
\vspace{-3.5mm}

\paragraph{Computing certified lower bound $\rho_{\text{cert}}$.}
Recall that the predicted class of input data $\xo$ is $c$ and let $t$ be a targeted class. Given the magnitude of largest input perturbation $\epsilon$, we can check if the output $\Ph{m}_c(\x) - \Ph{m}_t(\x) > 0$ by applying the global bounds derived in \eqref{eq:global_ub} and \eqref{eq:global_lb}. In other words, given an $\epsilon$, we will check the condition if $\eta_{c,L} - \eta_{t,U} > 0$. If the condition is true, we can increase $\epsilon$; otherwise decrease $\epsilon$. Thus, the largest certified lower bound can be attained by a bisection on $\epsilon$. Note that although there is an explicit $\epsilon$ term in \eqref{eq:global_ub} and \eqref{eq:global_lb}, they are \textit{not} a linear function in $\epsilon$ because all the intermediate bounds of $\Ph{r}$ depend on $\epsilon$. Fortunately, we can still find $\rho_{\text{cert}}$ numerically via the aforementioned bisection method. On the other hand, also note that the derivation of output bounds $\Ph{r}$ in each building block depend on the range $[\lwbnd{r-1},\upbnd{r-1}]$ of the building block input (say $\Ph{r-1}$), which we call the intermediate bounds. The value of intermediate bounds can be computed similarly by treating $\Ph{r-1}$ as the final output of the sub-network which consists of all building blocks before layer $r-1$ and deriving the corresponding $\A{0}_U, \A{0}_L, \B{0}_U, \B{0}_L$ in \eqref{eq:form_output}. Thus, all the intermediate bounds also have the same explicit forms as \eqref{eq:global_ub} and \eqref{eq:global_lb} but substituted by its corresponding $\A{0}_U, \A{0}_L, \B{0}_U, \B{0}_L$. 
\vspace{-3.5mm}

\paragraph{Discussions: Fast-Lin and CROWN are special cases of CNN-Cert.} \fastlin \cite{weng2018towards} and \crownName \cite{zhang2018crown} are  special cases of CNN-Cert. In  \fastlin, two linear bounds with the same slope (i.e. $\alpha_U = \alpha_L$ in \eqref{eq:conv_idea}) are applied on the ReLU activation while in \crownName and \cnncert different slopes are possible ($\alpha_U$ and $\alpha_L$ can be different). However, both \fastlin and \crownName only consider fully-connected layers (MLP) while \cnncert can handle various building blocks and architectures such as residual blocks, pooling blocks and batch normalization blocks and is hence a more general framework. We show in Table 13 (appendix) that when using the same linear bounds in ReLU activations, \cnncert obtains the same robustness certificate as \crownName; meanwhile, for the general activations, \cnncert uses more accurate linear bounds and thus achieves better certificate quality up to 260\% compared with \crownName (if we use exactly the same linear bounds, then \cnncert and \crownName indeed get the same certificate).  Note that in all cases, CNN-Cert is much faster than \crownName (2.5-11.4$\times$ speed-up) due to the advantage of explicit convolutional bounds in \cnncert. 

\vspace{-3.5mm}

\paragraph{Discussion: CNN-Cert is computationally efficient.}
\cnncert has a similar cost to forward-propagation for general convolutional neural networks -- it takes polynomial time, unlike algorithms that find the exact minimum adversarial distortion such as \reluplex~\cite{katz2017reluplex} which is NP-complete. As shown in the experiment sections, \cnncert demonstrates an empirical speedup as compared to (a) the original versions of \fastlin (b) an optimized sparse matrix versions of \fastlin (by us) and (c) Dual-LP approaches while maintaining similar or better certified bounds (the improvement is around 8-20 \%). For a pure CNN network with $m$ layers, $k$-by-$k$ filter size, $n$ filters per layer, input size $r$-by-$r$, and stride $1$-by-$1$, the time complexity of CNN-Cert is $O(r^2m^6k^4n^3)$. The equivalent fully connected network requires $O(r^6m^2n^3)$ time to certify.

\vspace{-3.5mm}

\paragraph{Discussion: Training-time operations are independent of \cnncert.} Since \cnncert is certifying the robustness of a fixed classifier $f$ at the testing time, techniques that only apply to the training phase, such as dropout, will not affect the operation of \cnncert (though the given model to be certified might vary if model weights differ).


\section{Experiments} \label{sec:simu}


\begin{table*}[t]
  \centering
  \caption{Averaged bounds of \cnncert and other methods on \textbf{(I) pure CNN networks with ReLU activations} , see \textbf{Comparative Methods} section for methods descriptions. '-' indicates the method is computationally infeasible.}
  \begin{adjustbox}{max width = 0.93\textwidth}
    \begin{tabular}{c|ccccc||cc||c||c}
    \hline
    Network & \multicolumn{5}{c||}{Certified lower bounds} & \multicolumn{2}{c||}{\cnncertada Improvement (\%)} & Attack & Uncertified \\
    \hline
          & $\ell_p$ norm & \cnncertada & \fastlin & Global-Lips & Dual-LP & vs. Fast-Lin & vs. Dual-LP & CW/EAD & CLEVER \\
    \hline
    MNIST, 4 layer & $\ell_\infty$ & 0.0491 & 0.0406 & 0.0002 & 0.0456 & +21\% & +8\%  & 0.1488 & 0.0542 \\
    5 filters & $\ell_2$ & 0.1793 & 0.1453 & 0.0491 & 0.1653 & +23\% & +8\%  & 3.1407 & 1.0355 \\
    8680 hidden nodes & $\ell_1$ & 0.3363 & 0.2764 & 0.0269 & 0.3121 & +22\% & +8\%  & 14.4516 & 4.2955 \\
    \hline
    MNIST, 4 layer & $\ell_\infty$ & 0.0340 & 0.0291 & 0.0000 & -     & +17\% & -     & 0.1494 & 0.0368 \\
    20 filters & $\ell_2$ & 0.1242 & 0.1039 & 0.0221 & -     & +20\% & -     & 3.0159 & 0.7067 \\
    34720 hidden nodes & $\ell_1$ & 0.2404 & 0.1993 & 0.0032 & -     & +21\% & -     & 13.7950 & 3.4716 \\
    \hline
    MNIST, 5 layer & $\ell_\infty$ & 0.0305 & 0.0248 & 0.0000 & -     & +23\% & -     & 0.1041 & 0.0576 \\
    5 filters & $\ell_2$ & 0.1262 & 0.1007 & 0.0235 & -     & +25\% & -     & 1.8443 & 0.9011 \\
    10680 hidden nodes & $\ell_1$ & 0.2482 & 0.2013 & 0.0049 & -     & +23\% & -     & 11.6711 & 3.5369 \\
    \hline
    CIFAR, 7 layer & $\ell_\infty$ & 0.0042 & 0.0036 & 0.0000 & -     & +17\% & -     & 0.0229 & 0.0110 \\
    5 filters & $\ell_2$ & 0.0340 & 0.0287 & 0.0023 & -     & +18\% & -     & 0.6612 & 0.3503 \\
    19100 hidden nodes & $\ell_1$ & 0.1009 & 0.0843 & 0.0001 & -     & +20\% & -     & 12.5444 & 1.2138 \\
    \hline
    CIFAR, 5 layer & $\ell_\infty$ & 0.0042 & 0.0037 & 0.0000 & -     & +14\% & -     & 0.0172 & 0.0075 \\
    10 filters & $\ell_2$ & 0.0324 & 0.0277 & 0.0042 & -     & +17\% & -     & 0.4177 & 0.2390 \\
    29360 hidden nodes & $\ell_1$ & 0.0953 & 0.0806 & 0.0005 & -     & +18\% & -     & 11.6536 & 1.5539 \\
    \hline
    \end{tabular}%
\end{adjustbox}
  \label{tab:addlabel}%
\end{table*}%

\begin{table*}
  \centering
  \caption{Averaged runtime of \cnncert and other methods on \textbf{(I) pure CNN networks with ReLU activations}}
\begin{adjustbox}{max width=0.93\textwidth}
    \begin{tabular}{c|ccccc||cccc}
    \hline
    Network & \multicolumn{5}{c||}{Average Computation Time (sec)} & \multicolumn{4}{c}{\cnncertada Speed-up} \\
    \hline
          & $\ell_p$ norm & \cnncertada & \fastlin & Global-Lips & Dual-LP & vs. Fast-Lin, sparse & vs. Fast-Lin & vs. Dual-LP & vs. CLEVER \\
    \hline
    MNIST, 4 layer & $\ell_\infty$ & 2.33  & 9.03  & 0.0001 & 853.20 & 1.9   & 3.9   & 366.1 & 31.4 \\
    5 filters & $\ell_2$ & 0.88  & 9.19  & 0.0001 & 236.30 & 5.0   & 10.5  & 270.1 & 83.0 \\
    8680 hidden nodes & $\ell_1$ & 0.86  & 8.98  & 0.0001 & 227.69 & 5.2   & 10.5  & 265.2 & 87.1 \\
    \hline
    MNIST, 4 layer & $\ell_\infty$ & 17.27 & 173.43 & 0.0001 & -     & 2.0   & 10.0  & -     & 11.2 \\
    20 filters & $\ell_2$ & 17.19 & 180.10 & 0.0002 & -     & 2.1   & 10.5  & -     & 11.4 \\
    34720 hidden nodes & $\ell_1$ & 17.35 & 179.63 & 0.0001 & -     & 2.1   & 10.4  & -     & 11.0 \\
    \hline
    MNIST, 5 layer & $\ell_\infty$ & 4.96  & 16.89 & 0.0001 & -     & 1.4   & 3.4   & -     & 19.0 \\
    5 filters & $\ell_2$ & 2.25  & 18.47 & 0.0001 & -     & 3.0   & 8.2   & -     & 46.8 \\
    10680 hidden nodes & $\ell_1$ & 2.32  & 16.70 & 0.0001 & -     & 3.0   & 7.2   & -     & 43.6 \\
    \hline
    CIFAR, 7 layer & $\ell_\infty$ & 15.11 & 78.04 & 0.0001 & -     & 1.5   & 5.2   & -     & 12.3 \\
    5 filters & $\ell_2$ & 16.11 & 73.08 & 0.0001 & -     & 1.4   & 4.5   & -     & 11.8 \\
    19100 hidden nodes & $\ell_1$ & 14.93 & 76.89 & 0.0001 & -     & 1.5   & 5.1   & -     & 12.9 \\
    \hline
    CIFAR, 5 layer & $\ell_\infty$ & 20.87 & 169.29 & 0.0001 & -     & 1.5   & 8.1   & -     & 8.0 \\
    10 filters & $\ell_2$ & 16.93 & 170.42 & 0.0002 & -     & 2.0   & 10.1  & -     & 9.2 \\
    29360 hidden nodes & $\ell_1$ & 17.07 & 168.30 & 0.0001 & -     & 1.9   & 9.9   & -     & 9.3 \\
    \hline
    \end{tabular}%
    \end{adjustbox}
  \label{tab:addlabel}%
\end{table*}%


\begin{table*}[h!]
  \centering
\caption{Averaged bounds and runtimes on \textbf{(II) general CNN networks with ReLU activations}. }
    \begin{adjustbox}{max width=0.93\textwidth}
    \begin{tabular}{c|cccc|c||c||c||ccc}
    \hline
    Network & \multicolumn{4}{c|}{Certified Bounds} & \cnncertada Imp. (\%) & Attack & Uncertified & \multicolumn{3}{c}{Average Computation Time (sec)} \\
    \hline
          & $\ell_p$ norm & \cnncertrelu & \cnncertada & Global-Lips & vs. \cnncertrelu & CW/EAD & CLEVER & \cnncertada & Global-Lips & CW/EAD \\
    \hline
    MNIST, LeNet & $\ell_\infty$ & 0.0113 & 0.0120 & 0.0002 & +6\%  & 0.1705 & 0.0714 & 9.54  & 0.0001 & 20.50 \\
          & $\ell_2$ & 0.0617 & 0.0654 & 0.0600 & +6\%  & 5.1327 & 1.2580 & 9.46  & 0.0001 & 5.56 \\
          & $\ell_1$ & 0.1688 & 0.1794 & 0.0023 & +6\%  & 21.6101 & 5.5241 & 9.45  & 0.0001 & 3.75 \\
    \hline
    MNIST, 7 layer & $\ell_\infty$ & 0.0068 & 0.0079 & 0.0000 & +16\% & 0.1902 & 0.1156 & 191.81 & 0.0001 & 41.13 \\
          & $\ell_2$ & 0.0277 & 0.0324 & 0.0073 & +17\% & 4.9397 & 1.7703 & 194.82 & 0.0007 & 10.83 \\
          & $\ell_1$ & 0.0542 & 0.0638 & 0.0000 & +18\% & 19.6854 & 6.8565 & 188.84 & 0.0001 & 6.31 \\
    \hline
    MNIST, LeNet & $\ell_\infty$ & 0.0234 & 0.0273 & 0.0001 & +17\% & 0.1240 & 0.1261 & 10.05 & 0.0001 & 36.08 \\
    No Pooling & $\ell_2$ & 0.1680 & 0.2051 & 0.0658 & +22\% & 3.7831 & 2.4130 & 10.76 & 0.0003 & 8.17 \\
          & $\ell_1$ & 0.5425 & 0.6655 & 0.0184 & +23\% & 22.2273 & 10.6149 & 11.63 & 0.0001 & 5.34 \\
    \hline
    MNIST, 4 layer & $\ell_\infty$ & 0.0083 & 0.0105 & 0.0011 & +26\% & 0.0785 & 0.0318 & 2.35  & 0.0001 & 30.49 \\
    5 filters & $\ell_2$ & 0.0270 & 0.0333 & 0.3023 & +23\% & 0.8678 & 0.6284 & 2.42  & 0.0002 & 8.26 \\
     Batch Norm & $\ell_1$ & 0.0485 & 0.0604 & 0.1053 & +25\% & 6.1088 & 2.4622 & 2.39  & 0.0001 & 5.46 \\
    \hline
    MNIST, 4 layer & $\ell_\infty$ & 0.0406 & 0.0492 & 0.0002 & +21\% & 0.1488 & 0.0536 & 1.66  & 0.0001 & 22.23 \\
    5 filters & $\ell_2$ & 0.1454 & 0.1794 & 0.0491 & +23\% & 3.1407 & 1.0283 & 1.31  & 0.0001 & 5.78 \\
          & $\ell_1$ & 0.2764 & 0.3363 & 0.0269 & +22\% & 14.4516 & 4.4930 & 1.49  & 0.0001 & 3.98 \\
    \hline
    Tiny ImageNet & $\ell_\infty$ & 0.0002 & 0.0003 & -     & +24\% & 0.4773 & 0.0056 & 5492.35 & -     & 257.06 \\
    7 layer & $\ell_2$ & 0.0012 & 0.0016 & -     & +29\% & -     & 0.4329 & 5344.49 & -     & - \\
          & $\ell_1$ & 0.0038 & 0.0048 & -     & +28\% & -     & 7.1665 & 5346.08 & -     & - \\
    \hline \hline
    MNIST, LeNet & $\ell_\infty$ & 0.0117 & 0.0124 & 0.0003 & +6\%  & 0.1737 & 0.0804 & 6.89  & 0.0001 & 38.76 \\
    & $\ell_2$ & 0.0638 & 0.0678 & 0.0672 & +6\%  & 5.1441 & 1.4599 & 6.85  & 0.0001 & 9.22 \\
    100 images & $\ell_1$ & 0.1750 & 0.1864 & 0.0027 & +7\%  & 22.7232 & 5.7677 & 6.91  & 0.0001 & 5.57 \\

    \hline
    MNIST, 4 layer & $\ell_\infty$ & 0.0416 & 0.0500 & 0.0002 & +20\% & 0.1515 & 0.0572 & 0.98  & 0.0001 & 40.02 \\
    5 filters & $\ell_2$ & 0.1483 & 0.1819 & 0.0516 & +23\% & 3.2258 & 1.0834 & 0.85  & 0.0001 & 8.93 \\
    100 images & $\ell_1$ & 0.2814 & 0.3409 & 0.0291 & +21\% & 14.7665 & 4.2765 & 0.83  & 0.0001 & 6.25 \\
    \hline
    \end{tabular}%
    \end{adjustbox}
  \label{tab:addlabel}%
\end{table*}%


\begin{table*}[h!]
  \centering
\caption{Averaged bounds and runtimes on \textbf{(III) ResNet with ReLU activations} .}
\begin{adjustbox}{max width=0.93\textwidth}
    \begin{tabular}{c|ccc||c||c||c||ccc}
    \hline
    Network & \multicolumn{3}{c||}{Certified Bounds} & \cnncertada Imp. (\%) & Attack & Uncertified & \multicolumn{3}{|c}{Average Computation Time (sec)} \\
    \hline
          & $\ell_p$ norm & \cnncertrelu & \cnncertada & vs. \cnncertrelu & CW/EAD & CLEVER & \cnncertrelu & \cnncertada & CW/EAD \\
    \hline
    MNIST, ResNet-2 & $\ell_\infty$ & 0.0183 & 0.0197 & +8\%  & 0.0348 & 0.0385 & 2.26  & 2.25  & 24.96 \\
          & $\ell_2$ & 0.0653 & 0.0739 & +13\% & 0.2892 & 0.7046 & 2.21  & 2.25  & 5.59 \\
          & $\ell_1$ & 0.1188 & 0.1333 & +12\% & 4.8225 & 2.2088 & 2.19  & 2.22  & 3.00 \\
    \hline
    MNIST, ResNet-3 & $\ell_\infty$ & 0.0179 & 0.0202 & +13\% & 0.0423 & 0.0501 & 10.39 & 10.04 & 32.82 \\
          & $\ell_2$ & 0.0767 & 0.0809 & +5\%  & 0.3884 & 1.0704 & 10.13 & 10.11 & 6.89 \\
          & $\ell_1$ & 0.1461 & 0.1514 & +4\%  & 5.9454 & 3.8978 & 10.20 & 10.15 & 4.22 \\
    \hline
    MNIST, ResNet-4 & $\ell_\infty$ & 0.0153 & 0.0166 & +8\%  & 0.0676 & 0.0455 & 28.66 & 28.18 & 35.13 \\
          & $\ell_2$ & 0.0614 & 0.0683 & +11\% & 1.0094 & 0.9621 & 28.43 & 28.20 & 7.89 \\
          & $\ell_1$ & 0.1012 & 0.1241 & +23\% & 9.1925 & 3.7999 & 27.81 & 28.53 & 5.34 \\
    \hline
    \end{tabular}%
    \end{adjustbox}
  \label{tab:addlabel}%
\end{table*}%


\begin{table*}
    \centering
\caption{Averaged bounds and runtimes on \textbf{(IV) general CNNs and ResNet with general activation functions}. 7-layer sigmoid network results are omitted due to poor test accuracy.}
\begin{adjustbox}{max width=0.93\textwidth}
    \begin{tabular}{c|cccccc||c||ccccc}
    \hline
    Network & \multicolumn{6}{c||}{Certified lower bounds}         & Uncertified & \multicolumn{5}{c}{Average Computation Time (sec)} \\
    \hline
          & $\ell_p$ norm & \cnncertrelu & \cnncertada & Sigmoid & Tanh  & Arctan & CLEVER & \cnncertrelu & \cnncertada & Sigmoid & Tanh  & Arctan \\
    \hline
    MNIST, Pure CNN & $\ell_\infty$ & 0.0203 & 0.0237 & 0.0841 & 0.0124 & 0.0109 & 0.0354 & 18.34 & 18.27 & 18.81 & 20.31 & 19.03 \\
    8-layer & $\ell_2$ & 0.0735 & 0.0877 & 0.3441 & 0.0735 & 0.0677 & 0.4268 & 18.25 & 18.22 & 18.83 & 19.70 & 19.05 \\
    5 filters & $\ell_1$ & 0.1284 & 0.1541 & 0.7319 & 0.1719 & 0.1692 & 1.2190 & 18.35 & 18.51 & 19.40 & 20.00 & 19.36 \\
    \hline
    MNIST, General CNN & $\ell_\infty$ & 0.0113 & 0.0120 & 0.0124 & 0.0170 & 0.0153 & 0.0714 & 9.71  & 9.54  & 9.55  & 9.66  & 9.37 \\
    LeNet & $\ell_2$ & 0.0617 & 0.0654 & 0.0616 & 0.1012 & 0.0912 & 1.2580 & 9.45  & 9.46  & 9.42  & 9.49  & 9.50 \\
          & $\ell_1$ & 0.1688 & 0.1794 & 0.1666 & 0.2744 & 0.2522 & 5.5241 & 9.44  & 9.45  & 9.59  & 9.69  & 9.86 \\
    \hline
    MNIST, General CNN & $\ell_\infty$ & 0.0068 & 0.0079 & -     & 0.0085 & 0.0079 & 0.1156 & 193.68 & 191.81 & -     & 191.26 & 195.08 \\
    7-layer & $\ell_2$ & 0.0277 & 0.0324 & -     & 0.0429 & 0.0386 & 1.7703 & 194.21 & 194.82 & -     & 193.85 & 194.81 \\
          & $\ell_1$ & 0.0542 & 0.0638 & -     & 0.0955 & 0.0845 & 6.8565 & 187.88 & 188.84 & -     & 188.83 & 188.79 \\
    \hline
    MNIST, ResNet-3 & $\ell_\infty$ & 0.0179 & 0.0202 & 0.0042 & 0.0058 & 0.0048 & 0.0501 & 10.39 & 10.04 & 10.08 & 10.39 & 10.26 \\
          & $\ell_2$ & 0.0767 & 0.0809 & 0.0147 & 0.0223 & 0.0156 & 1.0704 & 10.13 & 10.11 & 10.14 & 10.43 & 10.27 \\
          & $\ell_1$ & 0.1461 & 0.1514 & 0.0252 & 0.0399 & 0.0277 & 3.8978 & 10.20 & 10.15 & 10.40 & 10.84 & 10.69 \\
    \hline
    \end{tabular}%
    \end{adjustbox}
  \label{tab:addlabel}%
\end{table*}%


\begin{table*}[h!]
  \centering
  \caption{Averaged bounds and runtimes on \textbf{(V) small MLP networks}. }
  \begin{adjustbox}{max width=0.93\textwidth}
    \begin{tabular}{c|ccccc||cc||c||c||c}
    \hline
    Network & \multicolumn{5}{c||}{Certified Bounds} & \multicolumn{2}{c||}{\cnncertada Improvement (\%)} & Exact & Attack & Uncertified \\
    \hline
          & $\ell_p$ norm & \fastlin & \cnncertrelu & \cnncertada & Dual-LP & vs. Fast-Lin & vs. Dual-LP & Reluplex & CW/EAD & CLEVER \\
    \hline
    MNIST, 2 layer & $\ell_\infty$ & 0.0365 & 0.0365 & 0.0371 & 0.0372 & +2\%  & 0\%   & 0.0830 & 0.0871 & 0.0526 \\
    20 nodes & $\ell_2$ & 0.7754 & 0.7754 & 0.7892 & 0.9312 & +2\%  & -15\% & -     & 1.9008 & 1.1282 \\
    Fully Connected & $\ell_1$ & 5.3296 & 5.3252 & 5.4452 & 5.7583 & +2\%  & -5\%  & -     & 15.8649 & 7.8207 \\
    \hline
    MNIST, 3 layer & $\ell_\infty$ & 0.0297 & 0.0297 & 0.0305 & 0.0308 & +3\%  & -1\%  & -   & 0.0835 & 0.0489 \\
    20 nodes & $\ell_2$ & 0.6286 & 0.6289 & 0.6464 & 0.7179 & +3\%  & -10\% & -     & 2.3083 & 1.0214 \\
    Fully Connected & $\ell_1$ & 4.2631 & 4.2599 & 4.4258 & 4.5230 & +4\%  & -2\%  & -     & 15.9909 & 6.9988 \\
    \hline
    \end{tabular}%
    \end{adjustbox}
  \label{tab:addlabel}%
\end{table*}%

We conduct extensive experiments comparing \cnncert with 
other lower-bound based verification methods on 5 classes of networks: (I) pure CNNs; (II) general CNNs (ReLU) with pooling and batch normalization; (III) residual networks (ReLU); (IV) general CNNs and residual networks with non-ReLU activation functions; (V) small MLP models. Due to page constraints, we refer readers to the appendix for additional results. Our codes are available at \url{https://github.com/AkhilanB/CNN-Cert}. 

\noindent \textbf{Comparative Methods.} 
\begin{itemize}
    \item Certification algorithms: (i) {\small\fastlin} provides certificate on ReLU networks~\cite{weng2018towards}; (ii) {\small\textsf{Global-Lips}} provides certificate using global Lipschitz constant~\cite{szegedy2013intriguing}; (iii) {\small \textsf{Dual-LP}} solves dual problems of the LP formulation in ~\cite{kolter2017provable}, and is the best result that \cite{dvijotham2018dual} can achieve, although it might not be attainable due to the any-time property; (iv) {\small \textsf{Reluplex}} \cite{katz2017reluplex} obtains exact minimum distortion but is computationally expensive.
    \item Robustness estimation, Attack methods: (i) {\small \textsf{CLEVER}} \cite{weng2018evaluating} is a robustness estimation score without certification; (ii) {\small \textsf{CW/EAD}} are attack methods~\cite{carlini2017towards,chen2017ead}.
    \item Our methods: \cnncertrelu is \cnncert with the same linear bounds on ReLU used in \fastlin, while \cnncertada uses adaptive bounds all activation functions. CNNs are converted into equivalent MLP networks before evaluation for methods that only support MLP networks.
\end{itemize}


\noindent \textbf{Implementations, Models and Dataset.} \cnncert is implemented with Python (numpy with numba) and we also implement a version of \fastlin using sparse matrix multiplication for comparison with \cnncert since convolutional layers correspond to sparse weight matrices. Experiments are conducted on a AMD Zen server CPU. We evaluate \cnncert and other methods on CNN models trained on the MNIST, CIFAR-10 and tiny Imagenet datasets. All pure convolutional networks use 3-by-3 convolutions. The general 7-layer CNNs use two max pooling layers and uses 32 and 64 filters for two convolution layers each. LeNet uses a similar architecture to LeNet-5~\cite{lecun1998gradient}, with the no-pooling version applying the same convolutions over larger inputs. The residual networks (ResNet) evaluated use simple residual blocks with two convolutions per block and ResNet with $k$ residual blocks is denoted as ResNet-$k$. We evaluate all methods on 10 random test images and attack targets (in order to accommodate slow verification methods) and also 100 images results for some networks in Table 5. It shows that the results of average 100 images are similar to average 10 imagess. We train all models for 10 epochs and tune hyperparameters to optimize validation accuracy.
\vspace{-5mm}

\paragraph{Results (I): pure CNNs with ReLU activation.}
Table 3 demonstrates that \cnncert bounds consistently improve on \fastlin over network size. \cnncert also improves on Dual-LP. Attack results show that all certified methods leave a significant gap on the attack-based distortion bounds (i.e. upper bounds on the minimum distortions). Table 4 gives the runtimes of various methods and shows that \cnncert is faster than \fastlin, with over an order of magnitude speed-up for the smallest network. \cnncert is also faster than the sparse version of \fastlin. The runtime improvement of \cnncert decreases with network size. Notably, \cnncert is multiple orders of magnitude faster than the Dual-LP method. Global-Lips is an analytical bound, but it provides very loose lower bounds by merely using the product of layer weights as the Lipschitz constant. In contrast, \cnncert 
takes into account the network output at the neuron level and thus can certify significantly larger lower bounds, and is around 8-20 \%  larger compared to \fastlin and Dual-LP approaches. 

\vspace{-3mm}

\paragraph{Results (II), (III): general CNNs and ResNet with ReLU activation.}
Table 5 gives certified lower bounds for various general CNNs including networks with pooling layers and batch normalization. \cnncert improves upon \fastlin style ReLU bounds (\cnncertrelu). Interestingly, the LeNet style network without pooling layers has certified bounds much larger than the pooling version while the network with batch normalization has smaller certified bounds. These findings provide some new insights on uncovering the relation between certified robustness and network architecture, and \cnncert could potentially be  leveraged to search for more robust networks.
Table 6 gives ResNet results and shows \cnncert improves upon \fastlin.

\vspace{-3.5mm}

\paragraph{Results (IV): general CNNs and ResNet with general activations.} Table 7 computes certified lower bounds for networks with 4 different activation functions. Some sigmoid network results are omitted due to poor test set accuracy. We conclude that \cnncert can indeed efficiently find non-trivial lower bounds for all the tested activation functions and that computing certified lower bounds for general activation functions incurs no significant computational penalty.

\vspace{-3.5mm}

\paragraph{Results (V): Small MLP networks.} Table 8 shows results on small MNIST MLP with 20 nodes per layer. For the small 2-layer network, we are able to run Reluplex and compute minimum adversarial distortion. It can be seen that the gap between the certified lower bounds method here are all around 2 times while CLEVER and attack methods are close to Reluplex though without guarantees.     

\section{Conclusion and Future Work}
In this paper, we propose \cnncert, a general and efficient verification framework for certifying robustness of CNNs. By applying our proposed linear bounding technique on each building block, \cnncert can handle a wide variety of network architectures including convolution, pooling, batch normalization, residual blocks, as well as general activation functions. Extensive experimental results under four different classes of CNNs consistently validate the superiority of \cnncert over other methods in terms of its effectiveness in solving tighter non-trivial certified bounds and its run time efficiency.

\section{Acknowledgement}
Akhilan Boopathy, Tsui-Wei Weng and Luca Daniel are partially supported by MIT-IBM Watson AI Lab.  

{ \small
\bibliography{reference}
\bibliographystyle{aaai}
}

\onecolumn

\section*{Appendix}
 
\subsection{(a) Derivation of Act-Conv block: $\A{r}_U, \B{r}_U, \A{r}_L, \B{r}_L$}
\paragraph{Our goal.} We are going to show that the output $\Ph{r}$ in \eqref{eq:conv_relation} can be bounded as follows:
\begin{equation*}
   \A{r}_{L,\text{act}} * \Ph{r-1} + \B{r}_{L,\text{act}} \leq \Ph{r} \leq \A{r}_{U,\text{act}} * \Ph{r-1} + \B{r}_{U,\text{act}}
\end{equation*}
where $\A{r}_{U,\text{act}}, \A{r}_{L,\text{act}}, \B{r}_{U,\text{act}}, \B{r}_{L,\text{act}}$ are constant tensors related to weights $\W{r}$ and bias $\bias{r}$ as well as the corresponding parameters $\alpha_L,\alpha_U,\beta_L,\beta_U$ in the linear bounds of each neuron. 

\paragraph{Notations.} Below, we will use subscript $(x,y,z)$ to denote the location of $\Ph{r}(\x)$ and its corresponding weight filter is denoted as $\W{r}_{(x,y,z)}$. Meanwhile, we will use subscripts $(i,j,k)$ to denote the location in the weight filter.

\paragraph{Derivations of upper bounds.} By definition, the $(x,y,z)$-th output $\Ph{r}_{(x,y,z)}$ is a convolution of previous output $\sigma(\Ph{r-1})$ with its corresponding filter $\W{r}_{(x,y,z)}$:
\begin{align}
    \Ph{r}_{(x,y,z)} 
    &= \W{r}_{(x,y,z)} * \sigma(\Ph{r-1}) + \bias{r}_{(x,y,z)} \label{eq:conv:1} \\
    &= \sum_{i,j,k} \W{r}_{(x,y,z),(i,j,k)} \cdot [\sigma(\Ph{r-1})]_{(x+i,y+j,k)} + \bias{r}_{(x,y,z)} \label{eq:conv:2} \\
    & \leq \sum_{i,j,k} \W{r+}_{(x,y,z),(i,j,k)} \alpha_{U, (x+i,y+j,k)} (\Ph{r-1}_{(x+i,y+j,k)} + \beta_{U, (x+i,y+j,k)}) \nonumber \\ 
    &\quad + \W{r-}_{(x,y,z),(i,j,k)} \alpha_{L, (x+i,y+j,k)} (\Ph{r-1}_{(x+i,y+j,k)} + \beta_{L, (x+i,y+j,k)}) + \bias{r}_{(x,y,z)} \label{eq:conv:3} \\
    &= \A{r}_{U,(x,y,z)} * \Ph{r-1} + \B{r}_{U,(x,y,z)}. \label{eq:conv:4}
\end{align}
From \eqref{eq:conv:1} to \eqref{eq:conv:2}, we expand the convolution into summation form. From \eqref{eq:conv:2} to \eqref{eq:conv:3}, we apply the linear upper and lower bounds on each activation $[\sigma(\Ph{r-1})]_{(x+i,y+j,k)}$ as described in \eqref{eq:conv_idea}; the inequalities holds when multiplying with a positive weight and will be reversed (the RHS and LHS will be swapped) when multiplying with a negative weight. Since here we are deriving the upper bound, we only need to look at the RHS inequality. This is indeed the key idea for all the derivations. The tensor $\W{r+}_{(x,y,z)}$ contains only the positive entries of weights $\W{r}_{(x,y,z)}$ with all others set to zero while $\W{r-}_{(x,y,z)}$ contains only the negative entries of $\W{r}_{(x,y,z)}$ and sets other entries to zero. Note that with a slightly abuse of notation, the $\alpha_U,\alpha_L,\beta_U,\beta_L$ here are tensors with the same dimensions as $\Ph{r-1}$ (while the $\alpha_U,\alpha_L,\beta_U,\beta_L$ in \eqref{eq:conv_idea} are scalar), and we use subscripts to denote the entry of tensor, e.g. $\alpha_{U, (x+i,y+j,k)}$. 

From \eqref{eq:conv:3} to \eqref{eq:conv:4}, we define $\A{r}_{U,(x,y,z)}$ and $\B{r}_{U,(x,y,z)}$ as:
\begin{align}
    & \A{r}_{U,(x,y,z),(i,j,k)} 
    =  \W{r+}_{(x,y,z),(i,j,k)} \alpha_{U, (x+i,y+j,k)} + \W{r-}_{(x,y,z),(i,j,k)} \alpha_{L, (x+i,y+j,k)} \\
    &\B{r}_{U,(x,y,z)} 
    = \sum_{i,j,k} \W{r+}_{(x,y,z),(i,j,k)} \alpha_{U, (x+i,y+j,k)} \beta_{U, (x+i,y+j,k)} + \W{r-}_{(x,y,z),(i,j,k)} \alpha_{L, (x+i,y+j,k)} \beta_{L, (x+i,y+j,k)}
\end{align}
Note that $\B{r}_{U,(x,y,z)}$ can be written in the following convloution form simply by the definition of convolution:
\begin{align}
    \B{r}_{U,(x,y,z)} &= \W{r+}_{(x,y,z)}*(\alpha_U \odot \beta_U) + \W{r-}_{(x,y,z)}*(\alpha_L \odot \beta_L)
\end{align}
\paragraph{Derivations of lower bounds.}
The lower bounds can be derived similarly:
\begin{align}
    \Ph{r}_{(x,y,z)} 
    &= \W{r}_{(x,y,z)} * \sigma(\Ph{r-1}) + \bias{r}_{(x,y,z)} \nonumber \\
    &= \sum_{i,j,k} \W{r}_{(x,y,z),(i,j,k)} \cdot [\sigma(\Ph{r-1})]_{(x+i,y+j,k)} + \bias{r}_{(x,y,z)} \nonumber \\
    & \geq \sum_{i,j,k} \W{r+}_{(x,y,z),(i,j,k)} \alpha_{L, (x+i,y+j,k)} (\Ph{r-1}_{(x+i,y+j,k)} + \beta_{L, (x+i,y+j,k)}) \nonumber \\ 
    &\quad + \W{r-}_{(x,y,z),(i,j,k)} \alpha_{U, (x+i,y+j,k)} (\Ph{r-1}_{(x+i,y+j,k)} + \beta_{U, (x+i,y+j,k)}) + \bias{r}_{(x,y,z)} \nonumber \\
    &= \A{r}_{L,(x,y,z)} * \Ph{r-1} + \B{r}_{L,(x,y,z)}, \nonumber
\end{align}
where 
\begin{align}
    \A{r}_{L,(x,y,z)} &= \W{r+}_{(x,y,z),(i,j,k)} \alpha_{L, (x+i,y+j,k)} + \W{r-}_{(x,y,z),(i,j,k)} \alpha_{U, (x+i,y+j,k)}\\
    \B{r}_{L,(x,y,z),(i,j,k)} &= \W{r+}_{(x,y,z)}*(\alpha_L \odot \beta_L) + \W{r-}_{(x,y,z)}*(\alpha_U \odot \beta_U).
\end{align}

\clearpage

\subsection{(b) Derivation of Residual block: $\A{r}_U, \B{r}_U, \A{r}_L, \B{r}_L$}
\paragraph{Our goal.} We are going to show that the output $\Ph{r+2}$ in the residual block can be bounded as follows:
\begin{equation*}
   \A{r+2}_{L,\text{res}} * \Ph{r} + \B{r+2}_{L,\text{res}} \leq \Ph{r+2} \leq \A{r+2}_{U,\text{res}} * \Ph{r} + \B{r+2}_{U,\text{res}}
\end{equation*}
where $\Ph{r+2}$ denote the output of residual block (before activation), $\Ph{r+1}$ be the output of first convolutional layer and $\Ph{r}$ be the input of residual block, $\A{r+2}_{U,\text{res}}, \A{r+2}_{L,\text{res}}, \B{r+2}_{U,\text{res}}, \B{r+2}_{L,\text{res}}$ are constant tensors related to weights $\W{r+2}$, $\W{r+1}$, bias $\bias{r+2}$, $\bias{r+1}$, and the corresponding parameters $\alpha_L,\alpha_U,\beta_L,\beta_U$ in the linear bounds of each neuron. The input/output relation of residual block is as follows:
\begin{align}
    & \Ph{r+1} = \W{r+1}*\Ph{r} + \bias{r+1} \nonumber \\
    & \Ph{r+2} = \W{r+2}*\sigma(\Ph{r+1}) + \bias{r+2}+ \Ph{r} \nonumber
\end{align}

\paragraph{Notations.} Below, we will use subscript $(x,y,z)$ to denote the location of $\Ph{r}(\x)$ and its corresponding weight filter is denoted as $\W{r}_{(x,y,z)}$. Meanwhile, we will use subscripts $(i,j,k)$ to denote the location in the weight filter.

\paragraph{Derivations of upper bounds.} Write out $\Ph{r+2} = \W{r+2}*\sigma(\Ph{r+1}) + \bias{r+2}+ \Ph{r}$ and apply the act-conv bound on the term $\W{r+2}*\sigma(\Ph{r+1}) + \bias{r+2}$, we obtain 
\begin{equation*}
  \Ph{r+2} \leq \A{r+2}_{U,\text{act}} * \Ph{r+1} + \B{r+2}_{U,\text{act}}+ \Ph{r}.  
\end{equation*}
Plug in the equation $\Ph{r+1} = \W{r+1}*\Ph{r} + \bias{r+1}$, we get
\begin{align}
    \Ph{r+2} 
    & \leq \A{r+2}_{U,\text{act}} * (\W{r+1}*\Ph{r} + \bias{r+1}) + \B{r+2}_{U,\text{act}}+ \Ph{r} \nonumber\\
    & = (\A{r+2}_{U,\text{act}}*\W{r+1}) *\Ph{r} + \A{r+2}_{U,\text{act}} * \bias{r+1} + \B{r+2}_{U,\text{act}}+ \Ph{r}  \nonumber\\
    & = (\A{r+2}_{U,\text{act}}*\W{r+1} + I) *\Ph{r} +  \A{r+2}_{U,\text{act}} * \bias{r+1} + \B{r+2}_{U,\text{act}} \nonumber\\
    & = \A{r+2}_{U,\text{res}} * \Ph{r} + \B{r+2}_{U,\text{res}} \nonumber,
\end{align}
where  
\begin{align}
    & \A{r+2}_{U,\text{res}} = (\A{r+2}_{U,\text{act}}*\W{r+1} + I) \\
    & \B{r+2}_{U,\text{res}} = \A{r+2}_{U,\text{act}} * \bias{r+1} + \B{r+2}_{U,\text{act}}.
\end{align}
\paragraph{Derivations of lower bounds.} It is straight-forward to derive lower bounds following the above derivation of upper bound, and we have 
\begin{align}
    \Ph{r+2} 
    & \geq \A{r+2}_{L,\text{act}} * (\W{r+1}*\Ph{r} + \bias{r+1}) + \B{r+2}_{L,\text{act}}+ \Ph{r} \nonumber\\
    & = (\A{r+2}_{L,\text{act}}*\W{r+1} + I) *\Ph{r} +  \A{r+2}_{L,\text{act}} * \bias{r+1} + \B{r+2}_{L,\text{act}} \nonumber\\
    & = \A{r+2}_{L,\text{res}} * \Ph{r} + \B{r+2}_{L,\text{res}} \nonumber,
\end{align}
with 
\begin{align}
    & \A{r+2}_{L,\text{res}} = (\A{r+2}_{L,\text{act}}*\W{r+1} + I) \\
    & \B{r+2}_{L,\text{res}} = \A{r+2}_{L,\text{act}} * \bias{r+1} + \B{r+2}_{L,\text{act}}.
\end{align}

\clearpage

\subsection{(c) Derivation of Pooling block: $\A{r}_U, \B{r}_U, \A{r}_L, \B{r}_L$}
\paragraph{Our goal.} We are going to show that the output $\Ph{r}$ in the max-pooling block can be bounded as: 
\begin{equation}
    \A{r}_{L,\text{pool}} * \Ph{r-1} + \B{r}_{L,\text{pool}} \leq \Ph{r} \leq  \A{r}_{U,\text{pool}} * \Ph{r-1} + \B{r}_{U,\text{pool}}
\end{equation}
when the input $\Ph{r-1}$ is bounded in the range $[\lwbnd{r}, \upbnd{r}]$, and $\A{r}_{U,\text{pool}}, \A{r}_{L,\text{pool}}, \B{r}_{U,\text{pool}},\B{r}_{L,\text{pool}}$ are constant tensors related to $\lwbnd{r}$ and $\upbnd{r}$. The relationship between input and output is the following:  
\begin{equation}
    \Ph{r}_k = \max_{S_k} \; \Ph{r-1}_{S_k},
\end{equation}
where $S_k$ denotes the input index set (inputs that will be pooled) associated with the $k$-th output.

\paragraph{Notations.} We use index $1:n$ to denote the associated index in the pooling set $S_k$ for simplicity. 

\paragraph{Derivation of upper bound.} Suppose there are $n$ variables $x_1,x_2,...x_n$ with lower bounds $l_1,l_2,..l_n$ and upper bounds $u_1,u_2,...u_n$. Assume that all variables could be the maximum (if there exist some index $i$, $j$ s.t. $u_i \leq l_j $, then we can simply discard index $i$ since it can never be the maximum), so the maximum lower bound is lower than the minimum upper bound:
\begin{align}
    \max\{l_1, l_2, ...l_n\} < \min\{u_1, u_2, ... u_n\}
\end{align}
Define $\gamma_0=\frac{\sum_{i}{\frac{u_i}{u_i-l_i}-1}}{\sum_{i}{\frac{1}{u_i-l_i}}}$ and let
\begin{align}
\gamma = \min\{\max\{\gamma_0, \max\{l_1, l_2, ...l_n\}\}, \min\{u_1, u_2, ... u_n\}\},
\end{align}
we have $0 \leq \frac{u_i-\gamma}{u_i-l_i} \leq 1$. Then, consider the function 
\begin{align}
    U(x_1,x_2,...x_n) = \sum_{i}{\frac{u_i-\gamma}{u_i-l_i}(x_i-l_i)}+\gamma
\end{align}
Note that $U(l_1, l_2, ..., l_n)=\gamma$. Also, $U(l_1, l_2, ... l_{i-1}, u_i, l_{i+1}, ..., l_n)=u_i$. We show that $U(x_1,x_2,...,x_n)$ is an upper bound of $\max\{x_1,x_2,..., x_n\}$ by considering the following are three cases: (i) $\sum_{i}{\frac{x_i-l_i}{u_i-l_i}}=1$, (ii) $\sum_{i}{\frac{x_i-l_i}{u_i-l_i}}>1$, (iii) $\sum_{i}{\frac{x_i-l_i}{u_i-l_i}}<1$. 

\begin{itemize}
    \item Case (i): Note that when the $x_i$ are any of $(u_1, l_2, ..., l_n), (l_1, u_2, ..., l_n), ... ,(l_1, l_2, ..., u_n)$, then $\sum_{i}{\frac{x_i-l_i}{u_i-l_i}}=1$. Because $\sum_{i}{\frac{x_i-l_i}{u_i-l_i}}$ is linear in the $x_i$, $\sum_{i}{\frac{x_i-l_i}{u_i-l_i}}=1$ is a hyperplane containing $(u_1, l_2, ..., l_n), (l_1, u_2, ..., l_n), ... (l_1, l_2, ..., u_n)$. Note that any $(x_1, x_2, ... x_n)$ in the hyperplane $\sum_{i}{\frac{x_i-l_i}{u_i-l_i}}=1$ with $l_i\leq x_i \leq u_i$ lies inside the simplex with vertices $(u_1, l_2, ..., l_n),$ $(l_1, u_2, ..., l_n), ... (l_1, l_2, ..., u_n)$. Therefore, for any valid $(x_1, x_2, ... x_n)$ satisfying $\sum_{i}{\frac{x_i-l_i}{u_i-l_i}}=1$, the $x_i$ are a convex combination of $(u_1, l_2, ..., l_n), (l_1, u_2, ..., l_n), ..., (l_1, l_2, ..., u_n)$. Since the bound holds at $(u_1, l_2, ..., l_n), (l_1, u_2, ..., l_n), ..., (l_1, l_2, ..., u_n)$ and the maximum function is convex, the upper bound holds over the entire region $\sum_{i}{\frac{x_i-l_i}{u_i-l_i}}=1$.
    \item Case (ii): Consider reducing the values of the non-maximum $x_i$ by $\delta>0$. Set $\delta$ such that $\sum_{i}{\frac{x_i'-l_i}{u_i-l_i}}=1$ where $x_i'$ are the updated $x_i$. Since coefficients $\frac{u_i-\gamma}{u_i-l_i}\geq 0$, the bound decreases: $U(x_1, x_2, ...,x_n)\geq U(x_1', x_2', ...,x_n')$. However, the maximum does not change: $\max\{x_1,x_2,...,x_n\}=\max\{x_1',x_2',...,x_n'\}$. By case (i), the new bound is still an upper bound on the maximum:
        \begin{align}
        U(x_1', x_2', ...,x_n')\geq\max\{x_1',x_2',...,x_n'\}
        \end{align}
        Therefore the upper bound is valid in case (ii): $U(x_1, x_2, ...,x_n)\geq\max\{x_1,x_2,...,x_n\}$. 
    \item Case (iii): Consider increasing the value of the maximum $x_i$ by $\delta>0$ while keeping the other $x_i$ constant.  Set $\delta$ such that $\sum_{i}{\frac{x_i'-l_i}{u_i-l_i}}=1$ where $x_i'$ are the updated $x_i$. Since coefficients        $\frac{u_i-\gamma}{u_i-l_i}\leq 1$, the upper bound increases by a quantity less than $\delta$: $U(x_1, x_2, ..., x_n)+\delta\geq U(x_1', x_2', ..., x_n')$. The maximum increases by $\delta$:
        \begin{align}
        \max\{x_1,x_2,...,x_n\}+\delta=\max\{x_1',x_2',...,x_n'\}
        \end{align}
    By case (i), the new bound is still an upper bound on the maximum: $U(x_1', x_2', ...x_n')\geq\max\{x_1',x_2',...x_n'\}$. Substituting into the previous inequality: 
        \begin{align}
        U(x_1, x_2, ...x_n)+\delta\geq\max\{x_1,x_2,...x_n\}+\delta
        \end{align}
    Therefore, $U(x_1, x_2, ...x_n)\geq\max\{x_1,x_2,...x_n\}$. 
\end{itemize}

\noindent Thus, we have 
\begin{align}
    & \A{r}_{U,(x,y,z),(i,j,k)} = \frac{\upbnd{}_{(x+i,y+j,z)}-\gamma}{\upbnd{}_{(x+i,y+j,z)}-\lwbnd{}_{(x+i,y+j,z)}} \\
    & \B{r}_{U,(x,y,z),(i,j,k)} = \sum_{\vec{i} \in S_n} \frac{(\gamma-\upbnd{}_{(\vec{i}+\vec{x},z)})\lwbnd{}_{(\vec{i}+\vec{x},z)}}{\upbnd{}_{(\vec{i}+\vec{x},z)}-\lwbnd{}_{(\vec{i}+\vec{x},z)}} + \gamma
\end{align}

\paragraph{Derivation lower bound.}
The lower bound of $\max\{x_1,x_2,...x_n\}$ can be derived similarly as upper bound. First, define 
\begin{align}
    G = \sum_{i}{\frac{u_i-\gamma}{u_i-l_i}}
\end{align}
Note that if $G=1$, then $\gamma=\gamma_0$. If $G<1$, then $\gamma = \max\{l_1, l_2, ... l_n\}$. If $G>1$, then $\gamma = \min\{u_1, u_2, ... u_n\}$. Define 
\[
 \eta = 
  \begin{cases} 
   \min\{l_1, l_2, ..., l_n\} & \text{, if } G < 1 \\
   \max\{u_1, u_2, ..., u_n\} & \text{, if } G > 1 \\
   \gamma & \text{, if } G = 1
  \end{cases}
\]
and consider the function 
\begin{align}
L(x_1, x_2, ...x_n) = \sum_{i}{\frac{u_i-\gamma}{u_i-l_i}(x_i-\eta)+\eta}.
\end{align}
We show that $L$ is a lower bound on the maximum function by considering the following three cases (i) $G=1$, (ii) $G<1$, (iii) $G>1$:
\begin{itemize}
    \item Case (i): The bound reduces to $L(x_1, x_2, ...,x_n) = \sum_{i}{\frac{u_i-\gamma}{u_i-l_i}x_i}$. Since the $\frac{u_i-\gamma}{u_i-l_i} \geq 0$ and $\max\{x_1, x_2, ... x_n\}\geq x_i$:  
        \begin{align}
        L(x_1, x_2, ...,x_n) \leq \sum_{i}{\frac{u_i-\gamma}{u_i-l_i}}\max\{x_1,x_2,...,x_n\}=\max\{x_1,x_2,...,x_n\}
        \end{align}
    Therefore, the bound holds in case (u).
    \item Case (ii): The bound reduces to $ L(x_1, x_2, ...x_n) = \sum_{i}{\frac{u_i-\gamma}{u_i-l_i}x_i+\min\{l_1, l_2, ... l_n\}(1-G)}$. Since $\frac{u_i-\gamma}{u_i-l_i}\geq 0$ and $\max\{x_1, x_2, ... x_n\}\geq x_i$:
        \begin{align}
            L(x_1, x_2, ...x_n) \leq G\max\{x_1, x_2, ..., x_n\}+\min\{l_1, l_2, ..., l_n\}(1-G)
        \end{align}
    ,which can be expressed as:
        \begin{align}
            L(x_1, x_2, ...x_n) \leq G(\max\{x_1, x_2, ..., x_n\}-\min\{l_1, l_2, ..., l_n\})+\min\{l_1, l_2, ..., l_n\}
        \end{align}
    Since $G<1$ and $\max\{x_1, x_2, ... x_n\}-\min\{l_1, l_2, ... l_n\}\geq0$:
        \begin{align}
            L(x_1, x_2, ...,x_n) \leq \max\{x_1, x_2, ... ,x_n\}-\min\{l_1, l_2, ... ,l_n\}+\min\{l_1, l_2, ... ,l_n\}=\max\{x_1, x_2, ... ,x_n\}
        \end{align}
    Therefore, the bound holds in case (ii).
    \item Case (iii): The bound reduces to $L(x_1, x_2, ...,x_n) = \sum_{i}{\frac{u_i-\gamma}{u_i-l_i}x_i+\max\{u_1, u_2, ... u_n\}(1-G)}$. Since $\frac{u_i-\gamma}{u_i-l_i}\geq 0$ and $\max\{x_1, x_2, ..., x_n\}\geq x_i$:
        \begin{align}
        L(x_1, x_2, ...x_n) \leq G\max\{x_1, x_2, ..., x_n\}+\max\{u_1, u_2, ..., u_n\}(1-G)
        \end{align}
    , which can be expressed as:
        \begin{align}
            L(x_1, x_2, ...x_n) \leq G(\max\{x_1, x_2, ..., x_n\}-\max\{u_1, u_2, ..., u_n\})+\max\{u_1, u_2, ..., u_n\}
        \end{align}
    Since $G>1$ and $\max\{x_1, x_2, ... x_n\}-\max\{u_1, u_2, ... u_n\}\leq0$:
        \begin{align}
            L(x_1, x_2, ...,x_n) \leq \max\{x_1, x_2, ..., x_n\}-\max\{u_1, u_2, ..., u_n\}+\max\{u_1, u_2, ..., u_n\}=\max\{x_1, x_2, ..., x_n\}
        \end{align}
    Therefore, the bound holds in case (iii).
\end{itemize}

\noindent Thus, we have 
\begin{align}
    & \A{r}_{L,(x,y,z),(i,j,k)} = \frac{\upbnd{}_{(x+i,y+j,z)}-\gamma}{\upbnd{}_{(x+i,y+j,z)}-\lwbnd{}_{(x+i,y+j,z)}} \\
    & \B{r}_{L,(x,y,z),(i,j,k)} = \sum_{\vec{i} \in S_n} \frac{(\gamma-\upbnd{}_{(\vec{i}+\vec{x},z)})\eta}{\upbnd{}_{(\vec{i}+\vec{x},z)}-\lwbnd{}_{(\vec{i}+\vec{x},z)}} + \eta
\end{align}

\begin{table*}[t]
\centering
\caption{Expression of $\A{r}_{U}$ and $\B{r}_{U}$ in the case with general strides and padding. $\A{r}_{L}$ and $\B{r}_{L}$ have exactly the same form as $\A{r}_{U}$ and $\B{r}_{U}$ but with $U$ and $L$ swapped.}
\label{tab:ABexpression1}
\begin{adjustbox}{max width = \textwidth}
\begin{tabular}{l|c|c}
\hline
Blocks           & $\A{r}_{U,(\vec{x},z),(\vec{i},k)}$  & $\B{r}_{U}$   \\ \hline
(i) Act-Conv Block  & {\small $\W{r+}_{(\vec{x},z),(\vec{i},k)}\alpha_{U,(\vec{i}+\vec{s} \odot \vec{x} - \vec{p},k)} + \W{r-}_{(\vec{x},z),(\vec{i},k)}\alpha_{L,(\vec{i}+\vec{s} \odot \vec{x} - \vec{p},k)}$}  & {\small $\W{r+}*(\alpha_{U} \odot \beta_{U}) + \W{r-}*(\alpha_{L} \odot \beta_{L}) + \bias{r}$}  \\
(ii) Residual Block & {\small $[\A{r}_{U,\text{act}}*\W{r-1} + I]_{(\vec{i},k),(\vec{x},z)}$} &  {\small $\A{r}_{U,\text{act}}*\bias{r-1}+\B{r}_{U,\text{act}}$} \\
(iv) Pooling Block &  $\frac{\upbnd{}_{(\vec{i}+\vec{s} \odot \vec{x} - \vec{p},z)}-\gamma}{\upbnd{}_{(\vec{i}+\vec{s} \odot \vec{x} - \vec{p},z)}-\lwbnd{}_{(\vec{i}+\vec{s} \odot \vec{x} - \vec{p},z)}}$ & at location $(\vec{x},z)$: $\sum_{\vec{i} \in S_n} \frac{(\gamma-\upbnd{}_{(\vec{i}+\vec{s} \odot \vec{x} - \vec{p},z)})\lwbnd{}_{(\vec{i}+\vec{s} \odot \vec{x} - \vec{p},z)}}{\upbnd{}_{(\vec{i}+\vec{s} \odot \vec{x} - \vec{p},z)}-\lwbnd{}_{(\vec{i}+\vec{s} \odot \vec{x} - \vec{p},z)}} + \gamma$ \\ 
& {\small $\gamma = \min \{ \max \{ \gamma_0, \max \lwbnd{}_{S} \}, \min \upbnd{}_{S} \}$} & {\small $\gamma_0 = \frac{\sum_{S} \frac{\upbnd{}_{S}}{\upbnd{}_{S}-\lwbnd{}_{S}}-1}{\sum_{S} \frac{1}{\upbnd{}_{S}-\lwbnd{}_{S}}}$}  \\
\hline \hline
\multicolumn{3}{l}{{\small Note 1: $(\vec{i},k) = (i,j,k)$ denotes filter coordinate indices and $(\vec{x},z) = (x,y,z)$ denotes output tensor indices.}} \\
\multicolumn{3}{l}{{\small Note 2: $\A{r}_{U},\B{r}_{U},\W{},\alpha,\beta,\upbnd{},\lwbnd{}$ are all tensors. $\W{r+},\W{r-}$ contains only the positive, negative entries of $\W{r}$ with other entries equal 0.}} \\
\multicolumn{3}{l}{{\small Note 3: $\A{r}_{L},\B{r}_{L}$ for pooling block are slightly different. Please see Appendix (c) for details.}}\\
\multicolumn{3}{l}{{\small Note 4: $\vec{s}$ and $\vec{p}$ denote strides and padding respectively. Convolutions are taken using these values.}} \\
\hline
\end{tabular}
\end{adjustbox}
\end{table*}

\clearpage

\subsection{(d) Additional experiment results}

\begin{table*}[h!]
  \centering
  \caption{Additional results for certified bounds on \textbf{(I) pure networks with ReLU activations}. The corresponding runtimes are in Table 11. }
    \begin{tabular}{c|cccc||c||c}
    \hline
    Network & \multicolumn{4}{c||}{Certified Bounds} & \cnncertada Improvement (\%) & Attack \\
    \hline
          & $\ell_p$ norm & \cnncertada & \fastlin & Global-Lips & vs. Fast-Lin & CW/EAD \\
    \hline
    MNIST, 2 layer & $\ell_\infty$ & 0.0416 & 0.0363 & 0.0071 & +15\% & 0.1145 \\
    5 filters & $\ell_2$ & 0.1572 & 0.1384 & 0.2574 & +14\% & 2.6119 \\
    3380 hidden nodes & $\ell_1$ & 0.3266 & 0.2884 & 0.9868 & +13\% & 15.3137 \\
    \hline
    MNIST, 3 layer & $\ell_\infty$ & 0.0426 & 0.0381 & 0.0017 & +12\% & 0.1163 \\
    5 filters & $\ell_2$ & 0.2225 & 0.1925 & 0.1219 & +16\% & 2.6875 \\
    6260 hidden nodes & $\ell_1$ & 0.5061 & 0.4372 & 0.1698 & +16\% & 15.5915 \\
    \hline
    MNIST, 6 layer & $\ell_\infty$ & 0.0244 & 0.0195 & 0.0000 & +25\% & 0.1394 \\
    5 filters & $\ell_2$ & 0.0823 & 0.0664 & 0.0102 & +24\% & 2.4474 \\
    12300 hidden nodes & $\ell_1$ & 0.1517 & 0.1217 & 0.0006 & +25\% & 11.6729 \\
    \hline
    MNIST, 7 layer & $\ell_\infty$ & 0.0267 & 0.0228 & 0.0000 & +17\% & 0.1495 \\
    5 filters & $\ell_2$ & 0.1046 & 0.0872 & 0.0033 & +20\% & 2.8178 \\
    13580 hidden nodes & $\ell_1$ & 0.1913 & 0.1578 & 0.0001 & +21\% & 14.3392 \\
    \hline
    MNIST, 8 layer & $\ell_\infty$ & 0.0237 & 0.0203 & 0.0000 & +17\% & 0.1641 \\
    5 filters & $\ell_2$ & 0.0877 & 0.0734 & 0.0017 & +19\% & 3.0014 \\
    14560 hidden nodes & $\ell_1$ & 0.1540 & 0.1284 & 0.0000 & +20\% & 11.7247 \\
    \hline
    CIFAR, 5 layer & $\ell_\infty$ & 0.0070 & 0.0063 & 0.0000 & +11\% & 0.0241 \\
    5 filters & $\ell_2$ & 0.0574 & 0.0494 & 0.0168 & +16\% & 0.5903 \\
    14680 hidden nodes & $\ell_1$ & 0.1578 & 0.1348 & 0.0037 & +17\% & 15.7545 \\
    \hline
    CIFAR, 6 layer & $\ell_\infty$ & 0.0035 & 0.0031 & 0.0000 & +13\% & 0.0153 \\
    5 filters & $\ell_2$ & 0.0274 & 0.0231 & 0.0021 & +19\% & 0.2451 \\
    17100 hidden nodes & $\ell_1$ & 0.0775 & 0.0649 & 0.0001 & +19\% & 7.6853 \\
    \hline
    CIFAR, 8 layer & $\ell_\infty$ & 0.0029 & 0.0026 & 0.0000 & +12\% & 0.0150 \\
    5 filters & $\ell_2$ & 0.0270 & 0.0228 & 0.0008 & +18\% & 0.2770 \\
    20720 hidden nodes & $\ell_1$ & 0.0834 & 0.0698 & 0.0000 & +19\% & 6.3574 \\
    \hline
    MNIST, 4 layer & $\ell_\infty$ & 0.0355 & 0.0310 & 0.0001 & +15\% & 0.1333 \\
    10 filters & $\ell_2$ & 0.1586 & 0.1336 & 0.0422 & +19\% & 3.0030 \\
    17360 hidden nodes & $\ell_1$ & 0.3360 & 0.2818 & 0.0130 & +19\% & 14.8293 \\
    \hline
    MNIST, 8 layer & $\ell_\infty$ & 0.0218 & 0.0180 & 0.0000 & +21\% & 0.1566 \\
    10 filters & $\ell_2$ & 0.0884 & 0.0714 & 0.0006 & +24\% & 2.4015 \\
    29120 hidden nodes & $\ell_1$ & 0.1734 & 0.1394 & 0.0000 & +24\% & 11.5198 \\
    \hline
    CIFAR, 7 layer & $\ell_\infty$ & 0.0030 & 0.0026 & 0.0000 & +15\% & 0.0206 \\
    10 filters & $\ell_2$ & 0.0228 & 0.0189 & 0.0005 & +21\% & 0.4661 \\
    38200 hidden nodes & $\ell_1$ & 0.0635 & 0.0521 & 0.0000 & +22\% & 9.5752 \\
    \hline
    MNIST, 8 layer & $\ell_\infty$ & 0.0147 & 0.0112 & 0.0000 & +31\% & 0.1706 \\
    20 filters & $\ell_2$ & 0.0494 & 0.0365 & 0.0002 & +35\% & 2.4260 \\
    58240 hidden nodes & $\ell_1$ & 0.0912 & 0.0673 & 0.0000 & +36\% & 10.1088 \\
    \hline
    CIFAR, 5 layer & $\ell_\infty$ & 0.0037 & 0.0032 & 0.0000 & +16\% & 0.0199 \\
    20 filters & $\ell_2$ & 0.0250 & 0.0207 & 0.0016 & +21\% & 0.4150 \\
    58720 hidden nodes & $\ell_1$ & 0.0688 & 0.0569 & 0.0001 & +21\% & 12.6631 \\
    \hline
    CIFAR, 7 layer & $\ell_\infty$ & 0.0024 & 0.0020 & 0.0000 & +20\% & 0.0200 \\
    20 filters & $\ell_2$ & 0.0175 & 0.0142 & 0.0002 & +23\% & 0.3909 \\
    76400 hidden nodes & $\ell_1$ & 0.0504 & 0.0406 & 0.0000 & +24\% & 10.1112 \\
    \hline
    \end{tabular}%
  \label{tab:addlabel}%
\end{table*}%

\begin{table*}[h!]
  \centering
  \caption{Additional results for runtimes of certified bounds for \textbf{(I) pure networks with ReLU activations}. The corresponding certified bounds results are in Table 10.}
    \begin{tabular}{c|cccc||cc}
    \hline
    Network & \multicolumn{4}{c||}{Average Computation Time (sec)} & \multicolumn{2}{c}{\cnncertada Speed-up} \\
    \hline
          & $\ell_p$ norm & \cnncertada & \fastlin & Global-Lips & vs. Fast-Lin, sparse & vs. Fast-Lin \\
    \hline
    MNIST, 2 layer & $\ell_\infty$ & 0.06  & 0.78  & 0.0001 & 25.2  & 12.7 \\
    5 filters & $\ell_2$ & 0.06  & 0.71  & 0.0002 & 20.9  & 11.7 \\
    3380 hidden nodes & $\ell_1$ & 0.05  & 0.74  & 0.0001 & 25.1  & 14.2 \\
    \hline
    MNIST, 3 layer & $\ell_\infty$ & 0.25  & 4.02  & 0.0001 & 9.4   & 15.9 \\
    5 filters & $\ell_2$ & 0.25  & 3.99  & 0.0001 & 9.2   & 15.8 \\
    6260 hidden nodes & $\ell_1$ & 0.26  & 4.43  & 0.0001 & 9.6   & 17.2 \\
    \hline
    MNIST, 6 layer & $\ell_\infty$ & 4.71  & 25.91 & 0.0001 & 2.1   & 5.5 \\
    5 filters & $\ell_2$ & 4.68  & 26.18 & 0.0002 & 2.3   & 5.6 \\
    12300 hidden nodes & $\ell_1$ & 4.64  & 25.94 & 0.0001 & 2.2   & 5.6 \\
    \hline
    MNIST, 7 layer & $\ell_\infty$ & 8.23  & 32.38 & 0.0001 & 1.5   & 3.9 \\
    5 filters & $\ell_2$ & 8.24  & 33.36 & 0.0002 & 1.6   & 4.0 \\
    13580 hidden nodes & $\ell_1$ & 8.20  & 34.98 & 0.0001 & 1.6   & 4.3 \\
    \hline
    MNIST, 8 layer & $\ell_\infty$ & 13.67 & 45.92 & 0.0001 & 1.2   & 3.4 \\
    5 filters & $\ell_2$ & 14.14 & 47.44 & 0.0002 & 1.2   & 3.4 \\
    14560 hidden nodes & $\ell_1$ & 12.82 & 112.29 & 0.0009 & 1.6   & 8.8 \\
    \hline
    CIFAR, 5 layer & $\ell_\infty$ & 8.06  & 98.65 & 0.0001 & 3.6   & 12.2 \\
    5 filters & $\ell_2$ & 4.18  & 42.16 & 0.0001 & 2.8   & 10.1 \\
    14680 hidden nodes & $\ell_1$ & 4.17  & 39.65 & 0.0001 & 2.9   & 9.5 \\
    \hline
    CIFAR, 6 layer & $\ell_\infty$ & 8.49  & 56.91 & 0.0001 & 2.0   & 6.7 \\
    5 filters & $\ell_2$ & 8.51  & 52.42 & 0.0001 & 1.9   & 6.2 \\
    17100 hidden nodes & $\ell_1$ & 8.41  & 55.18 & 0.0001 & 2.0   & 6.6 \\
    \hline
    CIFAR, 8 layer & $\ell_\infty$ & 23.63 & 89.88 & 0.0001 & 1.3   & 3.8 \\
    5 filters & $\ell_2$ & 28.87 & 121.58 & 0.0001 & 1.0   & 4.2 \\
    20720 hidden nodes & $\ell_1$ & 23.63 & 121.44 & 0.0001 & 1.2   & 5.1 \\
    \hline
    MNIST, 4 layer & $\ell_\infty$ & 3.17  & -     & 0.0001 & 3.6   & - \\
    10 filters & $\ell_2$ & 3.20  & -     & 0.0002 & 3.5   & - \\
    17360 hidden nodes & $\ell_1$ & 3.16  & -     & 0.0001 & 3.6   & - \\
    \hline
    MNIST, 8 layer & $\ell_\infty$ & 61.74 & -     & 0.0001 & 0.9   & - \\
    10 filters & $\ell_2$ & 61.72 & -     & 0.0001 & 1.0   & - \\
    29120 hidden nodes & $\ell_1$ & 61.38 & -     & 0.0001 & 0.9   & - \\
    \hline
    CIFAR, 7 layer & $\ell_\infty$ & 67.23 & -     & 0.0001 & 1.1   & - \\
    10 filters & $\ell_2$ & 67.13 & -     & 0.0001 & 1.0   & - \\
    38200 hidden nodes & $\ell_1$ & 68.66 & -     & 0.0001 & 1.1   & - \\
    \hline
    MNIST, 8 layer & $\ell_\infty$ & 422.15 & -     & 0.0001 & 0.6   & - \\
    20 filters & $\ell_2$ & 422.78 & -     & 0.0001 & 0.6   & - \\
    58240 hidden nodes & $\ell_1$ & 421.54 & -     & 0.0001 & 0.6   & - \\
    \hline
    CIFAR, 5 layer & $\ell_\infty$ & 98.58 & -     & 0.0001 & 1.2   & - \\
    20 filters & $\ell_2$ & 98.25 & -     & 0.0002 & 1.1   & - \\
    58720 hidden nodes & $\ell_1$ & 98.98 & -     & 0.0001 & 1.2   & - \\
    \hline
    CIFAR, 7 layer & $\ell_\infty$ & 432.05 & -     & 0.0001 & 0.7   & - \\
    20 filters & $\ell_2$ & 430.64 & -     & 0.0002 & 0.7   & - \\
    76400 hidden nodes & $\ell_1$ & 430.87 & -     & 0.0001 & 0.7   & - \\
    \hline
    \end{tabular}%
  \label{tab:addlabel}%
\end{table*}%



\begin{table*}[h!]
\centering
\caption{Additional results for bounds and runtimes on \textbf{(IV) general CNNs and ResNet with general activation functions}. 7-layer sigmoid network results are omitted due to poor test accuracy.}
\begin{adjustbox}{max width=\textwidth}
\begin{tabular}{c|cccccc||ccccc}
\hline
Network & \multicolumn{6}{c||}{Certified lower bounds}         & \multicolumn{5}{c}{Average Computation Time (sec)} \\
\hline
      & $\ell_p$ norm & \cnncertrelu  & \cnncertada   & Sigmoid & Tanh  & Arctan & \cnncertrelu  & \cnncertada   & Sigmoid & Tanh  & Arctan \\
\hline
MNIST, Pure CNN & $\ell_\infty$ & 0.0406 & 0.0492 & 0.0654 & 0.0223 & 0.0218 & 1.23  & 1.27  & 1.32  & 1.68  & 1.37 \\
4-layer & $\ell_2$ & 0.1454 & 0.1794 & 0.3445 & 0.1518 & 0.1391 & 1.22  & 1.24  & 1.38  & 1.86  & 1.46 \\
5 filters & $\ell_1$ & 0.2764 & 0.3363 & 0.7643 & 0.3642 & 0.3349 & 1.24  & 1.23  & 1.38  & 1.84  & 1.43 \\
\hline
MNIST, Pure CNN & $\ell_\infty$ & 0.0063 & 0.0070 & 0.0193 & 0.0113 & 0.0091 & 5.66  & 5.70  & 6.04  & 6.24  & 6.12 \\
5-layer & $\ell_2$ & 0.0495 & 0.0574 & 0.1968 & 0.1160 & 0.1015 & 5.67  & 5.70  & 6.27  & 6.50  & 6.36 \\
5 filters & $\ell_1$ & 0.1348 & 0.1579 & 0.6592 & 0.3379 & 0.3233 & 5.59  & 5.62  & 6.29  & 6.65  & 6.40 \\
\hline
CIFAR, Pure CNN & $\ell_\infty$ & 0.0036 & 0.0042 & -     & 0.0067 & 0.0083 & 20.47 & 20.41 & -     & 21.46 & 20.90 \\
7-layer & $\ell_2$ & 0.0287 & 0.0340 & -     & 0.0670 & 0.0900 & 20.22 & 20.22 & -     & 21.46 & 21.15 \\
5 filters & $\ell_1$ & 0.0843 & 0.1009 & -     & 0.2166 & 0.2658 & 20.10 & 20.23 & -     & 21.50 & 21.28 \\
\hline
MNIST, ResNet-2 & $\ell_\infty$ & 0.0183 & 0.0197 & 0.0063 & 0.0274 & 0.0130 & 2.26  & 2.25  & 2.28  & 2.41  & 2.33 \\
      & $\ell_2$ & 0.0653 & 0.0739 & 0.0312 & 0.0724 & 0.0367 & 2.21  & 2.25  & 2.25  & 2.34  & 2.26 \\
      & $\ell_1$ & 0.1188 & 0.1333 & 0.0691 & 0.1262 & 0.0647 & 2.19  & 2.22  & 2.24  & 2.34  & 2.26 \\
\hline
MNIST, ResNet-4 & $\ell_\infty$ & 0.0153 & 0.0166 & 0.0049 & 0.0082 & 0.0085 & 28.66 & 28.18 & 28.26 & 28.33 & 28.47 \\
      & $\ell_2$ & 0.0614 & 0.0683 & 0.0242 & 0.0292 & 0.0267 & 28.43 & 28.20 & 28.26 & 28.49 & 28.28 \\
      & $\ell_1$ & 0.1012 & 0.1241 & 0.0517 & 0.0511 & 0.0458 & 27.81 & 28.53 & 28.61 & 29.03 & 28.35 \\
\hline
MNIST, ResNet-5 & $\ell_\infty$ & 0.0061 & 0.0062 & 0.0110 & 0.0081 & 0.0075 & 64.68 & 63.87 & 64.49 & 64.13 & 64.15 \\
      & $\ell_2$ & 0.0361 & 0.0283 & 0.0401 & 0.0224 & 0.0301 & 64.66 & 66.22 & 65.13 & 65.10 & 64.74 \\
      & $\ell_1$ & 0.0756 & 0.0525 & 0.0578 & 0.0371 & 0.0509 & 63.70 & 63.72 & 64.52 & 64.51 & 64.74 \\
\hline
\end{tabular}%
\end{adjustbox}
\label{tab:addlabel}%
\end{table*}

\begin{table*}[h!]
  \centering
  \caption{Additional results comparing CROWN and CNN-Cert with general activation functions. Note that for Sigmoid, Tanh and Arctan, \cnncertada use more accurate linear bounds on activation functions and thus achieve better verification bounds. }
    \begin{adjustbox}{max width=\textwidth}
    \begin{tabular}{c|ccc||c||cc||c}
    \hline
    Network & \multicolumn{3}{c||}{Certified Bounds} & \cnncertada Imp. (\%) & \multicolumn{2}{c||}{Average Computation Time (sec)} & \cnncertada Speed-up \\
    \hline
          & $\ell_p$ norm & \cnncertada & CROWN \cite{zhang2018crown} & vs. CROWN & \cnncertada & CROWN & vs. CROWN \\
    \hline
    MNIST, 8-layer & $\ell_\infty$ & 0.0203 & 0.0203 & 0\%   & 18.34 & 45.92 & 2.50 \\
    5 filters & $\ell_2$ & 0.0735 & 0.0734 & 0\%   & 18.25 & 47.44 & 2.60 \\
    ReLU  & $\ell_1$ & 0.1284 & 0.1284 & 0\%   & 18.35 & 112.29 & 6.12 \\
    \hline
    MNIST, 8-layer & $\ell_\infty$ & 0.0237 & 0.0237 & 0\%   & 18.27 & 208.21 & 11.40 \\
    5 filters & $\ell_2$ & 0.0877 & 0.0877 & 0\%   & 18.22 & 208.17 & 11.42 \\
    Ada   & $\ell_1$ & 0.1541 & 0.1540 & 0\%   & 18.51 & 126.59 & 6.84 \\
    \hline
    MNIST, 8-layer & $\ell_\infty$ & 0.0841 & 0.0827 & 2\%   & 18.81 & 186.71 & 9.93 \\
    5 filters & $\ell_2$ & 0.3441 & 0.3381 & 2\%   & 18.83 & 180.46 & 9.58 \\
    Sigmoid & $\ell_1$ & 0.7319 & 0.7185 & 2\%   & 19.40 & 202.25 & 10.43 \\
    \hline
    MNIST, 8-layer & $\ell_\infty$ & 0.0124 & 0.0051 & 146\% & 20.31 & 188.15 & 9.26 \\
    5 filters & $\ell_2$ & 0.0735 & 0.0215 & 242\% & 19.70 & 219.83 & 11.16 \\
    Tanh  & $\ell_1$ & 0.1719 & 0.0478 & 260\% & 20.00 & 182.73 & 9.14 \\
    \hline
    MNIST, 8-layer & $\ell_\infty$ & 0.0109 & 0.0067 & 62\%  & 19.03 & 210.42 & 11.06 \\
    5 filters & $\ell_2$ & 0.0677 & 0.0364 & 86\%  & 19.05 & 203.11 & 10.66 \\
    Arctan & $\ell_1$ & 0.1692 & 0.0801 & 111\% & 19.36 & 179.29 & 9.26 \\
    \hline
    \end{tabular}%
    \end{adjustbox}
  \label{tab:addlabel}%
\end{table*}%

\end{document}